\documentclass{article}

\usepackage[nonatbib,preprint]{neuripsdata2024}
\usepackage{pdflscape}
\usepackage[utf8]{inputenc} % allow utf-8 input
\usepackage[T1]{fontenc}    % use 8-bit T1 fonts
\usepackage[hyperfootnotes=false]{hyperref}       % hyperlinks
\usepackage{url}            % simple URL typesetting
\usepackage{booktabs}       % professional-quality tables
\usepackage{amsfonts}       % blackboard math symbols
\usepackage{nicefrac}       % compact symbols for 1/2, etc.
\usepackage{microtype}      % microtypography
\usepackage[table]{xcolor}         % colors
\usepackage{graphicx} 
\usepackage{subcaption}
\usepackage{multirow}
\usepackage{amsmath}
\usepackage[symbol]{footmisc}
\usepackage{caption} 
\usepackage{pifont}

\usepackage{tikz} %tikz
\usepackage{pgfplots}
\usepgfplotslibrary{groupplots}
\usepackage{wrapfig}
\usepackage{adjustbox}

\newcommand*{\belowrulesepcolor}[1]{% 
  \noalign{% 
    \kern-\belowrulesep 
    \begingroup 
      \color{#1}% 
      \hrule height\belowrulesep 
    \endgroup 
  }%
} 
\newcommand*{\aboverulesepcolor}[1]{% 
  \noalign{% 
    \begingroup 
      \color{#1}% 
      \hrule height\aboverulesep 
    \endgroup 
    \kern-\aboverulesep 
  }%
} 

\title{AlleNoise: large-scale text classification benchmark dataset with real-world label noise}

\author{%
Alicja Rączkowska\footnotemark[1] \quad Aleksandra Osowska-Kurczab\footnotemark[1] \quad Jacek Szczerbiński\footnotemark[1] \\  \textbf{Kalina Jasinska-Kobus\footnotemark[1]} \quad \textbf{Klaudia Nazarko\footnotemark[1]} \\
 Machine Learning Research \\
 Allegro.com \\
\texttt{\{alicja.raczkowska, aleksandra.kurczab, jacek.szczerbinski,}\\
\texttt{kalina.kobus, klaudia.nazarko\}@allegro.com}
}
\pgfplotsset{compat=1.18}

\begin{document}
\maketitle

\begin{abstract} %c abstract
  Label noise remains a challenge for training robust classification models. Most methods for mitigating label noise have been benchmarked using primarily datasets with synthetic noise. While the need for datasets with realistic noise distribution has partially been addressed by web-scraped benchmarks such as WebVision and Clothing1M, those benchmarks are restricted to the computer vision domain. With the growing importance of Transformer-based models, it is crucial to establish text classification benchmarks for learning with noisy labels. In this paper, we present \textit{AlleNoise}, a new curated text classification benchmark dataset with real-world instance-dependent label noise, containing over 500,000 examples across approximately 5,600 classes, complemented with a meaningful, hierarchical taxonomy of categories. The noise distribution comes from actual users of a major e-commerce marketplace, so it realistically reflects the semantics of human mistakes. In addition to the noisy labels, we provide human-verified clean labels, which help to get a deeper insight into the noise distribution, unlike web-scraped datasets typically used in the field. We demonstrate that a representative selection of established methods for learning with noisy labels is inadequate to handle such real-world noise. In addition, we show evidence that these algorithms do not alleviate excessive memorization. As such, with \textit{AlleNoise}, we set the bar high for the development of label noise methods that can handle real-world label noise in text classification tasks. The code and dataset are available for download at \url{https://github.com/allegro/AlleNoise}.
% The need for datasets with realistic noise distribution has partially been addressed by web-scraped benchmarks such as WebVision and Clothing1M. However, those benchmarks are restricted to the computer vision domain
\end{abstract}

\footnotetext[1]{Equal contribution}

\section{Introduction}
\label{section:introduction}

The problem of label noise poses a sizeable challenge for classification models~\cite{frenay_classification_2014,song_learning_2022}. With modern deep neural networks, due to their capacity, it is possible to memorize all labels in a given training dataset~\cite{rolnick_deep_2018}. This, effectively, leads to overfitting to noise if the training dataset contains noisy labels, which in turn reduces the generalization capability of such models~\cite{arpit_closer_2017,zhang_understanding_generalization_2017,zhang_understanding_generalization_sequel_2021}. 

Most previous works on training robust classifiers have focused on analyzing relatively simple cases of synthetic noise~\cite{jindal_learning_2017,patrini_making_2017}, either uniform (i.e. symmetric) or class-conditional (i.e. asymmetric). It is a common practice to evaluate these methods using popular datasets synthetically corrupted with label noise, such as MNIST~\cite{deng_mnist_2012}, ImageNet~\cite{deng_imagenet_2009}, CIFAR~\cite{krizhevsky_learning_2009} or SVHN~\cite{netzer_reading_2011}. However, synthetic noise is not indicative of realistic label noise and thus deciding to use noisy label methods based on such benchmarks can lead to unsatisfactory results in real-world machine learning practice. Moreover, it has been shown that these benchmark datasets are already noisy themselves~\cite{northcutt_pervasive_2021,liu_noise_text_2022}, so the study of strictly synthetic noise in such a context is intrinsically flawed.

Realistic label noise is instance dependent, i.e. the labeling mistakes are caused not simply by label ambiguity, but by input uncertainty as well~\cite{goldberger_training_2017}. This is an inescapable fact when human annotators are responsible for the labeling process~\cite{krishna_embracing_2016}. 
However, many existing approaches for mitigating instance-dependent noise have one drawback in common - they had to, in some capacity, artificially model the noise distribution due to the lack of existing benchmark datasets~\cite{nguyen_robust_2022,gu_instance-dependent_2021,chen_beyond_2020,xia_part-dependent_2020,algan_label_2020,berthon_confidence_2021}. In addition, most of the focus in the field has been put on image classification, but with the ever-increasing importance of Transformer-based~\cite{transformer} architectures, the problem of label noise affecting the fine-tuning of natural language processing models needs to be addressed as well. There are many benchmark datasets for text data classification~\cite{maas_imdb_2011,lin_dataset_2019,wang_glue_2019,bhatia_extreme_2016}, but none of them are meant for the study of label noise. In most cases, the actual level of noise in these datasets is unknown, so using them for benchmarking label noise methods is unfeasible.

Moreover, the datasets used in this research area usually contain relatively few labels. The maximum reported number of labels is 1000~\cite{li_webvision_2017}. As such, there is a glaring lack of a benchmark dataset for studying label noise that provides realistic real-world noise, a high number of labels and text data at the same time.

We see a need for a textual benchmark dataset that would provide realistic instance-dependent noise distribution with a known level of label noise, as well as a relatively large number of target classes, with both clean and noisy labels. To this end, in this paper we provide the following main contributions:
\begin{itemize}
    \item We introduce \textit{AlleNoise} - a benchmark dataset for multi-class text classification with real-world label noise. The dataset consists of 502,310 short texts (e-commerce product titles) belonging to 5,692 categories (taken from a real product assortment tree). It includes a noise level of 15\%, stemming from mislabeled data points. This amount of noise reflects the actual noise distribution in the data source (Allegro.com e-commerce platform). For each of the mislabeled data instances, the true category label was determined by human domain experts.
    \item We benchmark a comprehensive selection of well-established methods for classification with label noise against the real-world noise present in \textit{AlleNoise} and compare the results to synthetic label noise generated for the same dataset. We provide evidence that the selected methods fail to mitigate real-world label noise, even though they are very effective in alleviating synthetic label noise.
\end{itemize}

\begin{figure}
    \centering
    \captionsetup{labelfont=bf}
    \begin{subfigure}[t]{.3\textwidth}
      \centering
      \includegraphics[width=\textwidth]{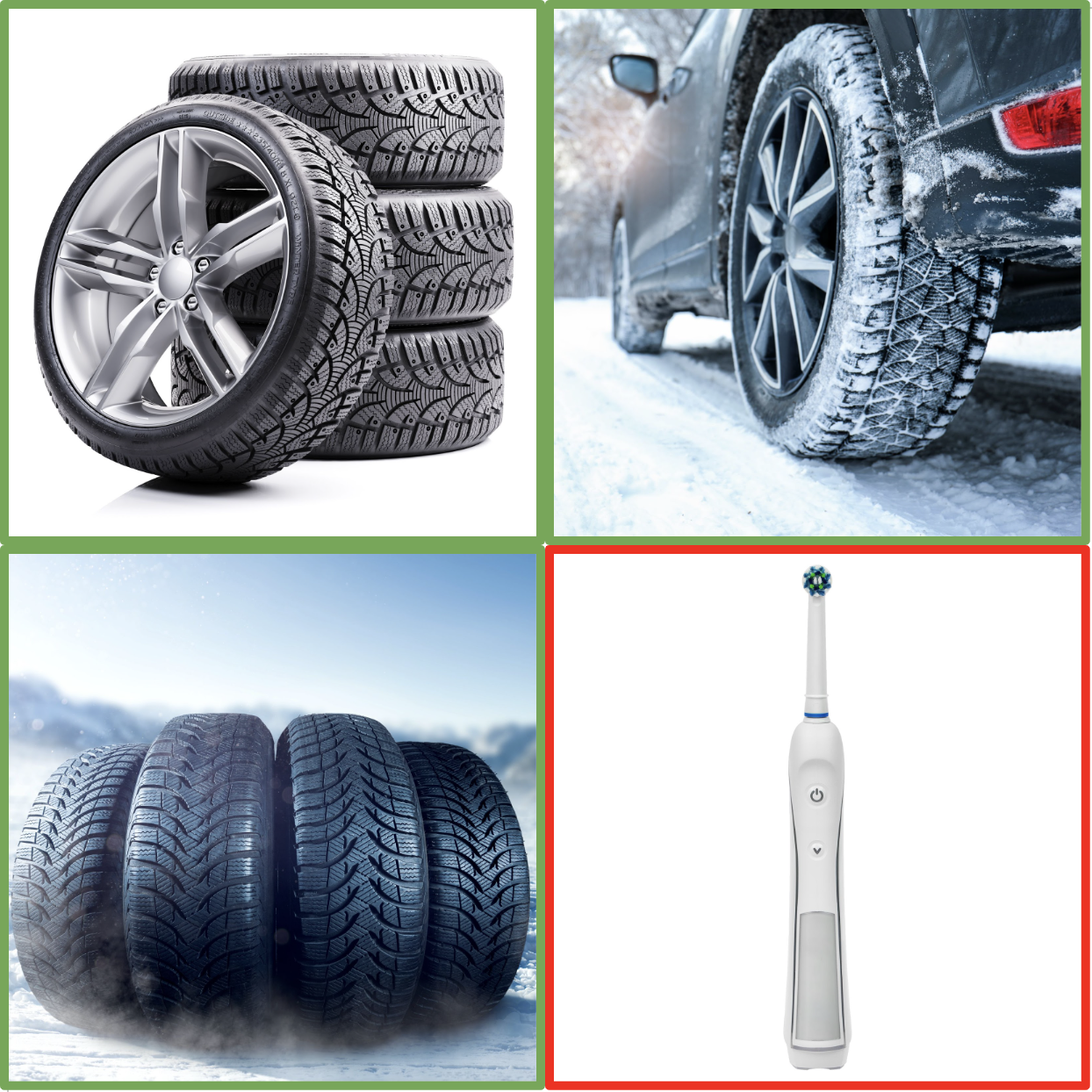}
      \caption{}
      \label{fig:sub1}
    \end{subfigure}%
    \hskip 20pt plus 0 fill
    \begin{subfigure}[t]{.3\textwidth}
      \centering
      \includegraphics[width=\textwidth]{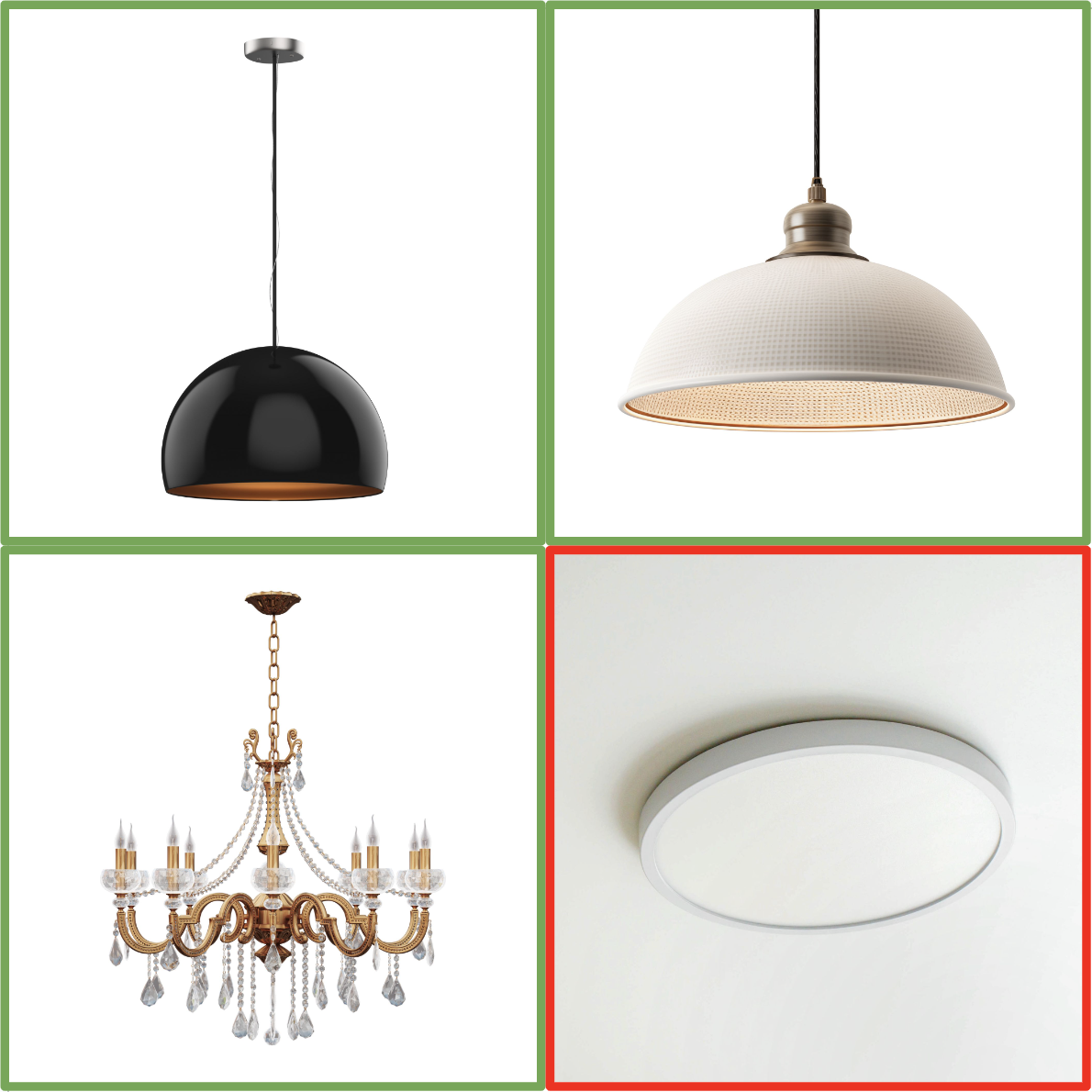}
      \caption{}
      \label{fig:sub2}
    \end{subfigure}
    \caption{
Symmetric noise vs. \textit{AlleNoise} in examples. Correct and noisy labels are marked in green and red, respectively. \textbf{(a)} Symmetric noise: an electric toothbrush incorrectly labeled as a winter tire is easy to spot, even for an untrained human. \textbf{(b)} \textit{AlleNoise}: a ceiling dome is mislabeled as a pendant lamp. This error is semantically challenging and hard to detect.
Note: \textit{AlleNoise} dataset does not include images.        
    }
    \label{fig:fig1-allenoise}
\end{figure}

% Depiction of the difference between symmetric noise and AlleNoise. Green boarder marks correctly labelled examples, red - incorrectly. Symmetric noise is unlikely to provide misleading examples. AlleNoise noisy examples are hard to detect even for untrained humans. AlleNoise dataset does not include images. We use images only to provide concise depiction of AlleNoise. 

\section{Related work}
\label{section:related_work}
% Several different types of algorithms have been developed for the purpose of combating the issue of label noise. One is to modify the loss function to become robust to label noise~\cite{kumar_self-paced_2010,liu_learning_2021,liu_early-learning_2020,englesson_generalized_2021, liu_peer_2020}. This modifies the training process so that noisy samples have less on an impact. Similarly, some researchers have pursued the direction of implicit regularization~\cite{han_co-teaching_2018,zhang_mixup_2018}, which also decreases the influence of noisy labels during the training. Then, it is also possible to directly filter out the noisy examples, either as a pre-processing step~\cite{northcutt_confident_2021, raczkowska_ara_2019} or during the training itself~\cite{kim_fine_2021}.

Several classification benchmarks with real-world instance-dependent noise have been reported in the literature. ANIMAL-10N~\cite{song_selfie_2019} is a human-labeled dataset of confusing images of animals, with 10 classes and an 8\% noise level. CIFAR-10N and CIFAR-100N~\cite{wei_learning_2022} are noisy versions of the CIFAR dataset, with labels assigned by crowd-sourced human annotators. CIFAR-10N is provided in three versions, with noise levels of 9\%, 18\% and 40\%, while CIFAR-100N has a noise level of 40\%. Clothing1M~\cite{xiao_learning_2015} is a large-scale dataset of fashion images crawled from several online shops. It contains 14 classes and the estimated noise rate is 38\%. Similarly, WebVision~\cite{li_webvision_2017} comprises of images crawled from the web, but it is more general - it has 1000 categories of diverse images. The estimated noise level is 20\%. DCIC~\cite{schmarje_annotation_2022} is a benchmark that consists of 10 real-world image datasets, with several human annotations per image. This allows for testing algorithms that utilize soft labels to mitigate various kinds of annotation errors. The maximum number of classes in the included datasets is 10.

\begin{table}[h]
    \centering
    \captionsetup{labelfont=bf}
    \begin{tabular}{cccccc}
        \toprule
        Dataset & Modality & Total examples & Classes & Noise level & Clean label \\ \midrule
        ANIMAL10N & Images & 55k & 10 & 8\% & \checkmark \\ %\midrule
        CIFAR10N & Images & 60k & 10 & 9/18/40\% & \checkmark \\% \midrule
        CIFAR100N & Images & 60k & 10 & 40\% & \checkmark \\ %\midrule
        WebVision & Images & 2.4M & 1000 & $\sim$20\% & \ding{55} \\% \midrule
        Clothing1M & Images & 1M & 14 & $\sim$38\% & \ding{55} \\% \midrule
        \midrule
        Hausa & Text & 2,917 & 5 & 50.37\% & \checkmark \\%\midrule
        Yor\`ub\'a & Text &  1,908 & 7 & 33.28\% & \checkmark \\% \midrule
        NoisyNER & Text & 217k & 4 & unspecified & \checkmark \\
        % \belowrulesepcolor{green!10}
        \rowcolor{green!10}
        \textbf{AlleNoise} & \textbf{Text} & \textbf{500k} & \textbf{5692} & \textbf{15\%} & \checkmark \\
        \hline
        % \aboverulesepcolor{green!10}
        % \bottomrule
    \end{tabular} 
    \caption{Comparison of \textit{AlleNoise} to previously published datasets created for studying the problem of learning with noisy labels. All datasets contain real-world noise. \textit{AlleNoise} is the biggest text classification dataset in this field, has a~known level of label noise and provides clean labels in addition to the noisy ones.}
    \label{tab:other_datasets}
\end{table}

With the focus in the label noise field being primarily on images, the issue of noisy text classification remains relatively unexplored. Previous works have either utilized existing classification datasets with synthetic noise~\cite{jindal_effective_2019,liu_noise_text_2022,nguyen_robust_2022} or introduced new datasets with real-world noise. NoisyNER~\cite{hedderich_analysing_2021} contains annotated named entity recognition data in the Estonian language, assigned to 4 categories. The authors do not mention the noise level, only that they provide 7 variants of real-world noise. NoisywikiHow~\cite{wu_noisywikihow_2023} is a dataset of articles scraped from the wikiHow website, with accompanying 158 article categories. The data was manually cleaned by human annotators, which eliminated the real-world noise distribution. The authors performed experiments by injecting synthetic noise into their dataset. Thus, NoisywikiHow is not directly comparable to \textit{AlleNoise}. Another two datasets are Hausa and Yor\`ub\'a~\cite{hedderich_transfer_2020}, text classification datasets of low-resource African languages with 5 and 7 categories respectively. They both include real-world noise with the level of 50.37\% for the former, and 33.28\% for the latter.

While there is a number of text datasets containing e-commerce product data~\cite{lin_dataset_2019,nguyen_robust_2022,bhatia_extreme_2016}, none of them have verified clean labels and in most cases the noise level is unknown. Similarly, classification settings with large numbers (i.e. more than 1000) of classes were not addressed up to this point in the existing datasets (\textbf{Tab. ~\ref{tab:other_datasets}}). 
% To the best of our knowledge, there are no established text classification datasets made for the express purpose of training models robust to real-world label noise (\textbf{Tab. ~\ref{tab:other_datasets}}).

% \section{Methods} % Ola
% \section{Learning with noisy labels} % Ola
% Let $\mathcal{X}$ denote the input feature space, and $\mathcal{Y}$ be a set of class labels. In a typical supervised setting, each instance $x_i$ has a true class label $y_i$. However, in learning with noisy labels, $\tilde{y}_i$ is observed instead, which is with an unknown probability $p$ (noise level) changed from the true $y_i$.

% In this setting, we train a classifier $f: \mathcal{X} \rightarrow \mathcal{Y}$ that generalizes knowledge learnt from a dataset $\mathcal{D}$, consisting of training examples $(x_i, \tilde{y}_i)$. Because $\tilde{y}_i$ can be affected by label noise, the model's predictions $\hat{y}_i = f(x_i)$ might be corrupted by the distribution of noisy labels as well. Maximizing the robustness of such a classifier implies reducing the impact of noisy training samples on the generalization performance.

\section{AlleNoise Dataset Construction}
\label{section:allenoise}
We introduce \textit{AlleNoise} - a benchmark dataset for large-scale multi-class text classification with real-world label noise.
The dataset consists of 502,310 e-commerce product titles listed on Allegro.com in 5,692 assortment categories, collected in January of 2022. 15\% of the products were listed in wrong categories, hence for each entry the dataset includes: 
% $x_i$ - the 
the product title, 
% $\tilde{y}_i$ - the 
the category where the product was originally listed, and 
% ${y}_i$ - the 
the category where it should be listed according to human experts. See Appendix \textbf{\ref{appendix:eda}} for exploratory data analysis of the dataset.

Additionally, we release the taxonomy of product categories in the form of a mapping (category~ID~$\rightarrow$~path~in~the~category~tree), which allows for fine-grained exploration of noise semantics. 

\begin{figure}
    \centering
    \captionsetup{labelfont=bf}
    % \begin{table}[]
\begin{tabular}{lll}
\toprule
% \multicolumn{1}{c}{Offer title} & \multicolumn{1}{c}{Noisy category label}   & \multicolumn{1}{c}{Clean category label}       \\ \midrule
\multicolumn{1}{c}{Offer title} & \multicolumn{1}{c}{Category label}   & \multicolumn{1}{c}{True category label}       \\ \midrule
Emporia PURE V25 BLACK                         & 352            & 170                 \\
Metal Hanging Lid Rack Suspended               & 68710          & 321104              \\
Miraculum Asta Plankton C Active Serum-Booster & 5360           & 89000  \\
\bottomrule
\end{tabular}
% \end{table}

    % \includegraphics[width=0.75\linewidth]{offers2.png}
    % \includegraphics[width=0.75\linewidth]{offers.png}
    \newline
    \newline
    
% \begin{table}[]
    \resizebox{\textwidth}{!}{%
\begin{tabular}{ll}
\toprule
\multicolumn{1}{c}{Category label} & \multicolumn{1}{c}{Category name}       \\ \midrule
352            & Electronics \textgreater~Phones and Accessories  \textgreater~GSM Accessories \textgreater~Batteries                    \\
170            & Electronics \textgreater~Phones and Accessories \textgreater~Smartphones and Cell Phones                                \\
68710          & Home and Garden \textgreater~Equipment \textgreater~Kitchen Utensils \textgreater~Pots and Pans \textgreater~Lids\\
321104         & Home and Garden \textgreater~Equipment \textgreater~Kitchen Utensils \textgreater~Pots and Pans \textgreater~Organizers \\
5360           & Allegro \textgreater~Beauty \textgreater~Care \textgreater~Face \textgreater~Masks                                      \\
89000          & Allegro \textgreater~Beauty \textgreater~Care \textgreater~Face \textgreater~Serum  \\
\bottomrule
\end{tabular}
}
% \end{table}
    % \includegraphics[width=1.0\linewidth]{categories2.png}
    % \includegraphics[width=1.0\linewidth]{categories.png}
    \caption{\textit{AlleNoise} consists of two tables: the first table includes the true and noisy label for each product title, while the second table maps the labels to category names.}
    \label{fig:dataset}
\end{figure}

% \subsubsection{Real-world noise}
\subsection{Real-world noise}
\label{section:real_world_noise}

% 74,094 mislabeled offers were collected from two sources:
We collected 74,094 mislabeled products from two sources: 1) customer complaints about a product being listed in the wrong category - such requests usually suggest the true category label, 2) assortment clean-up by internal domain experts, employed by Allegro - products listed in the wrong category were manually moved to the correct category.

The resulting distribution of label noise is not uniform over the entire product assortment - most of the noisy instances belong to a small number of categories. Such asymmetric distribution is an inherent feature of real-world label noise. It is frequently modeled with class-conditional synthetic noise in related literature. However, since the mistakes in \textit{AlleNoise} were based not only on the category name, but also on the product features, our noise distribution is in fact instance-dependent.

% \subsubsection{Clean data sampling}
\subsection{Clean data sampling}
\label{section:clean_data_sampling}

The 74,094 mislabeled products were complemented with 428,216 products listed in correct categories. The clean instances were sampled from the most popular items listed in the same categories as the noisy instances, proportionally to the total number of products listed in each category. The high popularity of the sampled products guarantees their correct categorization, because items that generate a lot of traffic are curated by human domain experts. Thus, the sampled distribution was representative for a subset of the whole marketplace: 5,692 categories out of over 23,000, for which label noise is particularly well known and described.

% \subsubsection{Post-processing}

\subsection{Post-processing}
\label{section:post_processing}
We automatically translated all 500k product titles from Polish to English. Machine translation is a common part of e-commerce, many platforms incorporate it in multiple aspects of their operation~\cite{tan_ecommerce_2020,zhang_improve_2023}. Moreover, it is an established practice to publish machine-translated text in product datasets~\cite{ni_justifying_2019}. Categories related to sexually explicit content were removed from the dataset altogether. Finally, categories with less than 5 products were removed from the dataset to allow for five-fold cross-validation in our experiments.
 
% \subsubsection{Synthetic noise generation}
\section{Methods}
% The following section describes how we benchmarked selected methods for training classifiers under label noise against \textit{AlleNoise} and how we compared the results with synthetic label noise generated for the same dataset.

\subsection{Problem statement}
\label{section:problem_statement}
Let $\mathcal{X}$ denote the input feature space, and $\mathcal{Y}$ be a set of class labels. In a typical supervised setting, each instance $x_i$ has a true class label $y_i$. However, in learning with noisy labels, $\tilde{y}_i$ is observed instead, which is with an unknown probability $p$ (noise level) changed from the true $y_i$.

In this setting, we train a classifier $f: \mathcal{X} \rightarrow \mathcal{Y}$ that generalizes knowledge learnt from a dataset $\mathcal{D}$, consisting of training examples $(x_i, \tilde{y}_i)$. Because $\tilde{y}_i$ can be affected by label noise, the model's predictions $\hat{y}_i = f(x_i)$ might be corrupted by the distribution of noisy labels as well. Maximizing the robustness of such a classifier implies reducing the impact of noisy training samples on the generalization performance. In the \textit{AlleNoise} dataset, $x_i$ corresponds to the product title, $\tilde{y}_i$ is the original product category, and $y_i$ is the correct category.

\subsection{Synthetic noise generation}
\label{section:synthetic_noise_generation}
In order to compare the real-world noise directly with synthetic noise, we applied different kinds of synthetic noise to the clean version of \textit{AlleNoise}: the synthetic noise was applied to each instance's true label ${y}_i$, yielding a new synthetic noisy label $\tilde{y}_i$. Overall, the labels were flipped for a controlled fraction $p = 15\%$ of all instances. 
% (15\% or 40\%).
We examined the following types of synthetic noise:
\begin{itemize}
    \item Symmetric noise: each instance is given a noisy label different from the original label, with uniform probability $p$.
    \item Class-conditional pair-flip noise: each instance in class $j$ is given a noisy label $j+1$ with probability $p$.
    \item Class-conditional nested-flip noise: we only flip categories that are close to each other in the hierarchical taxonomy of categories. For example, for the parent category \textit{Car Tires} we perform a cyclic flip between its children categories: \textit{Summer} $\rightarrow$ \textit{Winter} $\rightarrow$ \textit{All-Season} $\rightarrow$ \textit{Summer} with probability $p$. Thus, the noise transition matrix is a block matrix with a small number of off-diagonal elements equal to $p$.
    \item  Class-conditional matrix-flip noise: the transition matrix between classes is approximated with the baseline classifier's confusion matrix. The confusion matrix is evaluated against the clean labels on 8\% of the dataset (validation split)~\cite{patrini_making_2017}. The resulting noise distribution is particularly tricky: we flip the labels between the classes that the model is most likely to confuse.
\end{itemize}

\subsection{Model architecture} 
\label{setion:model_architecture}
Next, we evaluated several algorithms for training classifiers under label noise. For a fair comparison, all experiments utilized the same classifier architecture as well as training and evaluation loops. We followed a fine-tuning routine that is typical for text classification tasks. In particular, we vectorized text inputs with XLMRoberta~\cite{xlmroberta}, a multilingual text encoder based on the Transformer architecture~\cite{transformer}. To provide the final class predictions, we used a single fully connected layer with a softmax activation and the number of neurons equal to the number of classes. The baseline model uses cross-entropy (CE) as a loss function.

Models were trained with the AdamW optimiser and linear LambdaLR scheduling ($\text{warmup steps}=100$). We have not used any additional regularization, i.e. weight decay or dropout. Key training parameters, such as batch size ($\text{bs}=256$) and learning rate ($\text{lr}=10^{-4}$) were tuned to maximize the validation accuracy on the clean dataset. All models have been trained for 10 epochs. Training of the baseline model, accelerated with a single NVIDIA A100 40GB GPU, lasted for about 1 hour.

We used five-fold stratified cross-validation to comprehensively evaluate the results of the models trained with label noise. For each fold, the full dataset was divided into three splits: $\mathcal{D}_{train}$, $\mathcal{D}_{val}$, $\mathcal{D}_{test}$, in proportion 72\% : 8\% : 20\%. Following the literature on learning with noisy labels~\cite{song_learning_2022}, both $\mathcal{D}_{train}$ and $\mathcal{D}_{val}$ were corrupted with label noise, while $\mathcal{D}_{test}$ remained clean. 

All of the results presented in this study correspond to the last checkpoint of the model. We use the following format for presenting the experimental results: $[\text{m}] \pm [\text{s}]$, where $m$ is an average over the five cross-validation folds, while $s$ is the standard deviation. Experiments used a seeded random number generator to ensure the reproducibility of the results. 

\subsection{Evaluation metrics}
\label{section:metrics}
% \subsubsection{Evaluation metrics}
Accuracy on the clean test set is the key metric in our study. We expect that methods that are robust to the label noise observed in the training phase, should be able to improve the test accuracy when compared to the baseline model. 

Additionally, to better understand the difference between synthetic and real-world noise, we collected detailed validation metrics. The validation dataset $\mathcal{D}_{val}$ contained both instances for which the observed label $\tilde{y}_i$ was incorrect ($\mathcal{D}_{val}^{\texttt{noisy}}$) and correct ($\mathcal{D}_{val}^{\texttt{clean}}$). Noisy observations from $\mathcal{D}_{val}^{\texttt{noisy}}$ were used to measure the memorization metric $\texttt{memorized}_{val}$, defined as a ratio of predictions $\hat{y}_i$ that match the noisy label $\tilde{y}_i$. Notice that our memorization metric is computed on the validation set, contrary to the training set typically used in the literature~\cite{liu_early-learning_2020}. Our metric increases when the model not only memorizes incorrect classes from the training observations, but also repeats these errors on unseen observations. Furthermore, we compute accuracy on $\mathcal{D}_{val}^{\texttt{noisy}}$ denoted as $\texttt{correct}_{val}^{\texttt{noisy}}$ and its counterpart on the clean fraction, $\texttt{correct}_{val}^{\texttt{clean}}$.

\subsection{Benchmarked methods}
\label{section:benchmarked_methods}
We evaluated the following methods for learning with noisy labels: Self-Paced Learning (SPL)~\cite{kumar_self-paced_2010}, Provably Robust Learning (PRL)~\cite{liu_learning_2021}, Early Learning Regularization (ELR)~\cite{liu_early-learning_2020}, Generalized Jensen-Shannon Divergence (GJSD)~\cite{englesson_generalized_2021}, Co-teaching (CT)~\cite{han_co-teaching_2018}, Co-teaching+ (CT+)~\cite{yu_how_2019}, Mixup (MU)~\cite{zhang_mixup_2018} and Pseudo-Label Selection (PLS)~\cite{albert_noisecorrection_pls_2022}. Additionally, we implemented Clipped Cross-Entropy as a simple baseline (see Appendix~\textbf{A}). These approaches represent a comprehensive selection of different method families: novel loss functions (GJSD), noise filtration (SPL, PRL, CCE, CT, CT+), robust regularization (ELR, PLS), data augmentation (MU) and training loop modifications (CT, CT+, PLS).  

These methods are implemented with a range of technologies and software libraries. As such, in order to have a reliable and unbiased framework for comparing them, it is necessary to standardize the software implementation. To this end, we re-implemented these methods using PyTorch (version 1.13.1) and PyTorch Lightning (version 1.5.0) software libraries. We publish our re-implementations and the accompanying evaluation code on GitHub at \url{https://github.com/allegro/AlleNoise}.

To select the best hyperparameters (see Appendix~\textbf{A}) for each of the benchmarked algorithms, we performed a tuning process on the \textit{AlleNoise} dataset. We focused on maximizing the fraction of correct clean examples $\texttt{correct}_{val}^{\texttt{clean}}$ within the validation set for two noise types: 15\% real-world noise and 15\% symmetric noise. The tuning was performed on a single fold selected out of five cross-validation folds, yielding optimal hyperparameter values (\textbf{Tab.~S1}). We then used these tuned values in all further experiments. 

\section{Results}
\label{section:results}
The selected methods for learning with noisy labels were found to perform differently on AlleNoise than on several types of synthetic noise. Below we highlight those differences in performance and relate them to the dissimilarities between real-world and synthetic noise.

% Performance of the selected methods for learning with noisy labels, as benchmarked with the \textit{AlleNoise} real-world noise, differs from their performance on synthetic noise types. We measure their accuracy and analyse the impact of noise type on their performance, to clearly show the difference between real-world and synthetic noise.

\subsection{Synthetic noise vs \textit{AlleNoise}}
\label{section:synthetic_vs_allenoise}

The selected methods were compared on the clean dataset, the four types of synthetic noise and on the real-world noise in \textit{AlleNoise} (\textbf{Tab.~\ref{tab-pretrained-noise15}}). The accuracy score on the clean dataset did not \textcolor{black}{degrade} for any of the evaluated algorithms when compared to the baseline CE. When it comes to the performance on the datasets with symmetric noise, the best method was GJSD, with CCE not too far behind. GJSD increased the accuracy by 1.31 percentage points (p.p.) over the baseline. For asymmetric noise types, the best method was consistently ELR. It significantly improved the test accuracy in comparison to CE, by 1.3 p.p. on average. Interestingly, some methods deteriorated the test accuracy. CT+ was worse than the baseline for all synthetic noise types (by 2.59 p.p., 2.12 p.p., 3.1 p.p., 2.02 p.p. for symmetric, pair-flip, nested-flip and matrix-flip noises, respectively), while SPL decreased the results for all types of asymmetric noise (by 3.63 p.p., 4.2 p.p., 5.17 p.p. for pair-flip, nested-flip and matrix-flip noises, respectively). CT+ seems to perform better for noise levels higher than 15\% (see Appendix~\textbf{B}). On \textit{AlleNoise}, we observed nearly no improvement in accuracy for any of the evaluated algorithms, and CT+, PRL and SPL all deteriorated the metric (by 2.65 p.p., 2.05 p.p. and 4.61 p.p., respectively). 

\begin{table}[h]
\centering
\captionsetup{labelfont=bf}
\resizebox{\textwidth}{!}{%

\definecolor{mplcrimson}{RGB}{220,20,60}
\definecolor{mplpurple}{RGB}{128,0,128}
\definecolor{mplviolet}{RGB}{238,130,238}
\definecolor{mpllimegreen}{RGB}{50,205,50}
\definecolor{mplteal}{RGB}{0,128,128}
\definecolor{mplaqua}{RGB}{0,255,255}
\definecolor{mplblue}{RGB}{0,0,255}
\definecolor{mplcoral}{RGB}{255,127,80}
\definecolor{mplgold}{RGB}{255,215,0}
\definecolor{mplmagenta}{RGB}{255,0,255}
\definecolor{mpldarkmagenta}{RGB}{139,0,139}

\centering
\begin{tabular}{lr|rrrr|r}
\toprule
% \multicolumn{1}{c}{\multirow{3}{*}{}} & \multicolumn{1}{c}{\multirow{3}{*}{Clean set}} & \multicolumn{4}{c}{Synthetic 15\%} & \multicolumn{1}{c}{\multirow{3}{*}{\textbf{AlleNoise}}} \\
% \multicolumn{1}{c}{} & \multicolumn{1}{c}{} & \multicolumn{1}{c}{Symmetric} & \multicolumn{1}{c}{\begin{tabular}[c]{@{}c@{}}Asymmetric\\ pair-flip\end{tabular}} & \multicolumn{1}{c}{\begin{tabular}[c]{@{}c@{}}Asymmetric\\ nested-flip\end{tabular}} & \multicolumn{1}{c}{\begin{tabular}[c]{@{}c@{}}Asymmetric\\ matrix-flip\end{tabular}} & \multicolumn{1}{c}{} \\
% \multicolumn{1}{c}{} & \multicolumn{1}{c|}{Clean set} & \multicolumn{1}{c}{Symmetric} & \multicolumn{1}{c}{\begin{tabular}[c]{@{}c@{}}Asymmetric\\ pair-flip\end{tabular}} & \multicolumn{1}{c}{\begin{tabular}[c]{@{}c@{}}Asymmetric\\ nested-flip\end{tabular}} & \multicolumn{1}{c|}{\begin{tabular}[c]{@{}c@{}}Asymmetric\\ matrix-flip\end{tabular}} & \multicolumn{1}{c}{\textbf{AlleNoise}} \\
% \multicolumn{1}{c}{} & \multicolumn{1}{c|}{Clean set} & \multicolumn{1}{c}{Symmetric} & \multicolumn{1}{c}{\begin{tabular}[c]{@{}c@{}}Asymmetric\\ pair-flip\end{tabular}} & \multicolumn{1}{c}{\begin{tabular}[c]{@{}c@{}}Asymmetric\\ nested-flip\end{tabular}} & \multicolumn{1}{c|}{\begin{tabular}[c]{@{}c@{}}Asymmetric\\ matrix-flip\end{tabular}} & \multicolumn{1}{c}{\textbf{AlleNoise}} \\
   & \multicolumn{1}{c|}{Clean set} & \multicolumn{1}{c}{Symmetric} & \multicolumn{1}{c}{Pair-flip} & \multicolumn{1}{c}{Nested-flip} & \multicolumn{1}{c|}{Matrix-flip} & \multicolumn{1}{c}{AlleNoise} \\

\midrule
\textcolor{black}{CE} & \textbf{74.85 ± 0.15} & 71.97 ± 0.08 & 71.92 ± 0.08 & 71.77 ± 0.08 & 70.75 ± 0.17 & {63.71 ± 0.11} \\ \midrule
\textcolor{black}{ELR} & 74.81 ± 0.11 & 72.15 ± 0.10 & \textbf{73.21 ± 0.21} & \textbf{73.07 ± 0.11} & \textbf{72.02 ± 0.17} & {63.72 ± 0.19} \\% \midrule
\textcolor{black}{MU} & 74.73 ± 0.09 & 71.96 ± 0.08 & 71.95 ± 0.10 & 71.65 ± 0.14 & 70.73 ± 0.17 & 63.65 ± 0.12 \\ %\midrule
\textcolor{black}{CCE} & 74.80 ± 0.09 & 73.01 ± 0.10 & 71.86 ± 0.17 & 71.62 ± 0.10 & 70.61 ± 0.10 & \textbf{63.73 ± 0.22} \\ %\midrule
CT & *74.85 ± 0.15 & 72.42 ± 0.13 & 71.99 ± 0.14 & 71.55 ± 0.08 & 70.57 ± 0.18 & 63.32 ± 0.25 \\ %\midrule
CT+ & *74.85 ± 0.15 & $\downarrow$69.38 ± 0.29 & $\downarrow$69.80 ± 0.24 & $\downarrow$68.67 ± 2.59 & $\downarrow$68.73 ± 0.27 & $\downarrow$61.06 ± 0.38 \\ %\midrule
PRL & *74.85 ± 0.15 & 71.82 ± 0.17 & 71.95 ± 0.15 & 71.73 ± 0.16 & 71.12 ± 0.10 & $\downarrow$61.66 ± 0.17 \\ %\midrule
SPL & *74.85 ± 0.15 & 72.56 ± 0.10 & $\downarrow$68.29 ± 0.15 & $\downarrow$67.57 ± 0.14 & $\downarrow$65.58 ± 0.15 & $\downarrow$59.10 ± 0.14 \\ %\midrule
GJSD & 74.63 ± 0.10 & \textbf{73.28 ± 0.13} & 71.67 ± 0.15 & 71.40 ± 0.10 & 70.55 ± 0.17    & 63.63 ± 0.19 \\ 
PLS & 74.00 ± 0.16 & 71.31 ± 0.14 & 70.09 ± 0.15 & 70.02 ± 0.12 & 69.68 ± 0.18 & 63.07 ± 0.29 \\
\bottomrule
\end{tabular}

}
\caption{Accuracy of the evaluated methods on the clean dataset compared to various noisy datasets with 15\% noise level. The noisy datasets include \textit{AlleNoise}, symmetric synthetic noise, and asymmetric synthetic noises: pair-flip, nested-flip, and matrix-flip. * marks cases equivalent to the baseline CE. $\downarrow$ marks results significantly worse than the baseline CE. Best results for each noise type are bolded.}
\label{tab-pretrained-noise15}
\end{table}

\subsection{Noise type impacts memorization} 
\label{section:memorization}

% \begin{figure}[h]
%     \captionsetup{labelfont=bf}
%     \centering
%     \includegraphics[width=\textwidth]{figures-results/memorized_and_correct_15.png}
%     \caption{Memorization and correctness metrics as a function of the training step. \textbf{(a)} The value of $\texttt{memorized}_{val}$ for synthetic noise types. \textbf{(b)} The value of $\texttt{memorized}_{val}$ for \textit{AlleNoise}. \textbf{(c)} The value of $\texttt{correct}_{val}^{\texttt{clean}}$ for \textit{AlleNoise}. \textbf{(d)} The value of $\texttt{correct}_{val}^{\texttt{noisy}}$ for \textit{AlleNoise}.}
%     \label{fig:memorization}
% \end{figure}

\fboxsep=0mm%padding thickness
\fboxrule=1pt%border thickness

\def\imagetop#1{\vtop{\null\hbox{#1}}}
\begin{figure}
    \captionsetup{labelfont=bf}
    \begin{tabular}{@{\hskip2pt}l@{\hskip1pt}@{\hskip1pt}l@{\hskip1pt}@{\hskip1pt}l@{\hskip1pt}@{\hskip1pt}l@{\hskip1pt}@{\hskip1pt}l@{\hskip1pt}@{\hskip1pt}l@{\hskip1pt}}
    \imagetop{\textbf{(a)}} 
    & \multicolumn{5}{l}{\imagetop{
    \includegraphics[width=0.96\textwidth]{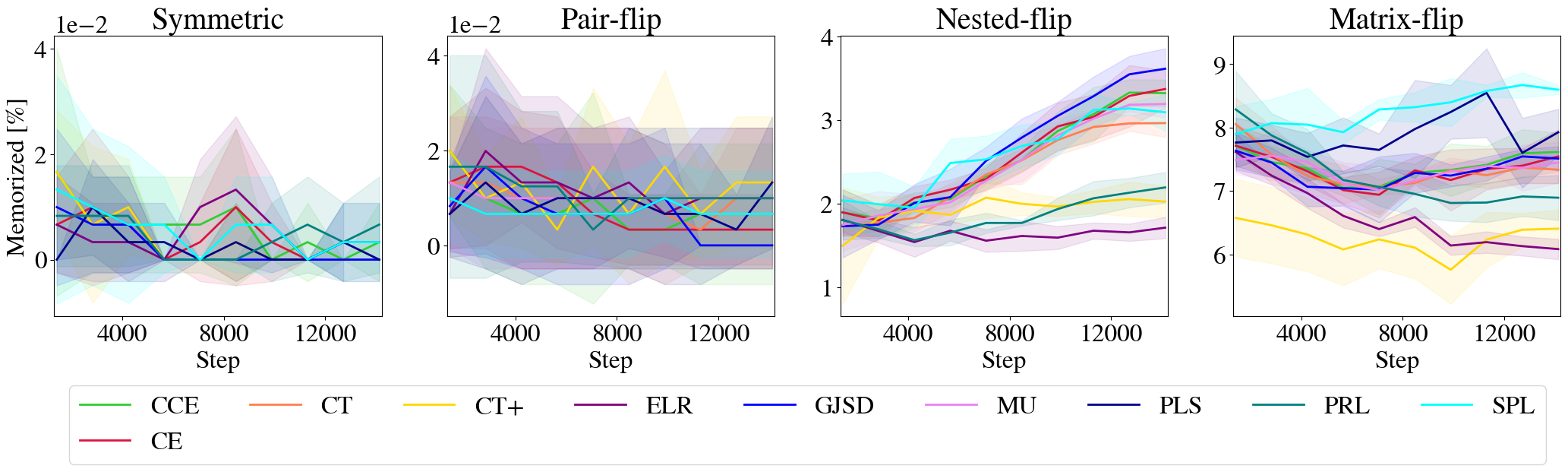}
    }}\\
    
    \imagetop{\textbf{(b)}} & \imagetop{
    \includegraphics[width=0.3\textwidth]{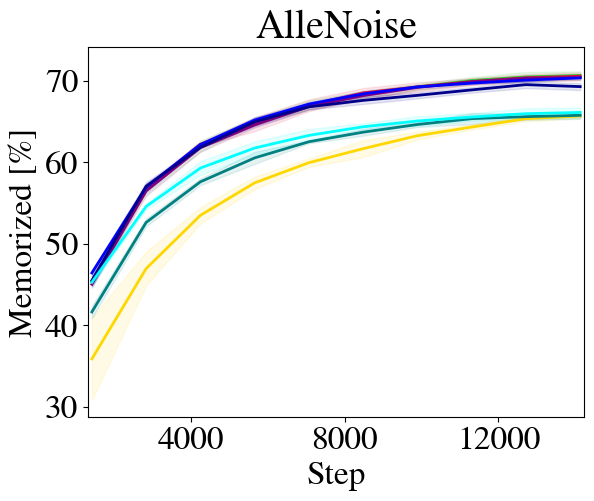}
    } &
    \imagetop{\textbf{(c)}} & \imagetop{
    \includegraphics[width=0.3\textwidth]{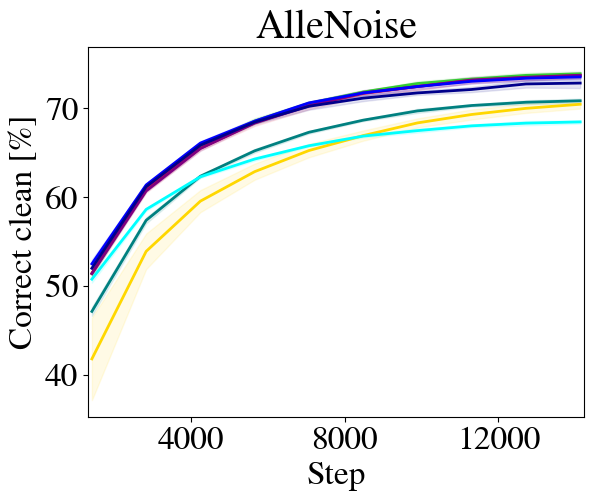}
    } &
    \imagetop{\textbf{(d)}} & \imagetop{
    \includegraphics[width=0.3\textwidth]{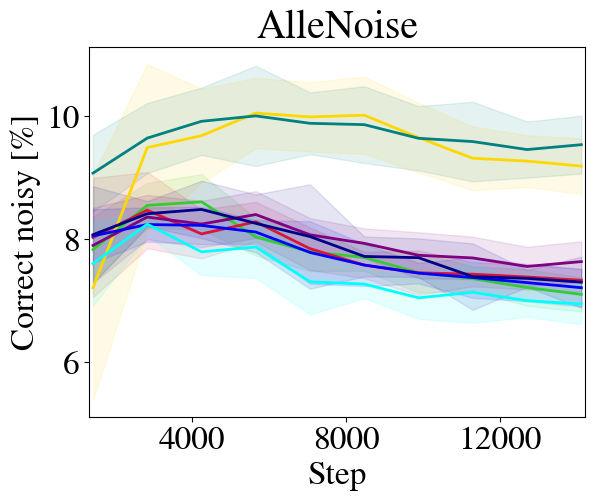} }
    \end{tabular}
    \caption{Memorization and correctness metrics as a function of the training step. \textbf{(a)} The value of $\texttt{memorized}_{val}$ for synthetic noise types. \textbf{(b)} The value of $\texttt{memorized}_{val}$ for \textit{AlleNoise}. \textbf{(c)} The value of $\texttt{correct}_{val}^{\texttt{clean}}$ for \textit{AlleNoise}. \textbf{(d)} The value of $\texttt{correct}_{val}^{\texttt{noisy}}$ for \textit{AlleNoise}.}
    \label{fig:memorization}
\end{figure}

To better understand the difference between synthetic noise types and \textit{AlleNoise}, we analyze how the $\texttt{memorized}_{val}^{\texttt{noisy}}$, $\texttt{correct}_{val}^{\texttt{noisy}}$ and $\texttt{correct}_{val}^{\texttt{clean}}$ metrics (see~\ref{section:metrics}) evolve over time. Memorization and correctness should be interpreted jointly with test accuracy (\textbf{Tab.~\ref{tab-pretrained-noise15}}). Additional plots of training, validation and test accuracy enriching this analysis are included in Appendix~\textbf{\ref{appendix:accuracy}}.

Synthetic noise types are memorized to a smaller extent than the real-world \textit{AlleNoise} (\textbf{Fig.~\ref{fig:memorization}a}). 
For the two simplest synthetic noise types, symmetric and pair-flip, the value of $\texttt{memorized}_{val}$ is negligible (very close to zero).
For the other two synthetic noise types, nested-flip and matrix-flip, memorization is still low (2-8\%), but there are clearly visible differences between the benchmarked methods.
While ELR, CT+ and PRL all keep the value of $\texttt{memorized}_{val}^{\texttt{noisy}}$ low for both nested-flip and matrix-flip noise types, it is only ELR that achieves test accuracy higher than the baseline. 

However, for \textit{AlleNoise}, the situation is completely different. All the training methods display increasing $\texttt{memorized}_{val}$ values throughout the training, up to 70\% (\textbf{Fig.~\ref{fig:memorization}b}). PRL, SPL and CT+ give lower memorization than the other methods, but this is not reflected in higher accuracy. While these methods correct some of the errors on noisy examples, as measured by $\texttt{correct}_{val}^{\texttt{noisy}}$ (\textbf{Fig.~\ref{fig:memorization}d}), they display $\texttt{correct}_{val}^{\texttt{clean}}$ lower than other tested approaches (\textbf{Fig.~\ref{fig:memorization}c}), and thus overall they achieve low accuracy.

% In general, the methods based on filtering out the noisy instances performed poorly. Noise filtration is only effective if the filtering criterion is well suited to the noise distribution. Otherwise, it leads to losing both clean and noisy training instances and a deterioration of model accuracy.

These results show that reducing memorization is necessary to create noise-robust classifiers. In this context, it is clear that \textit{AlleNoise}, with its real-world instance-dependent noise distribution, is a challenge for the existing methods.

\subsection{Noise distribution}
\label{section:noise_distribution}
\begin{figure}[h]
    \centering
    \captionsetup{labelfont=bf}
    \includegraphics[width=0.95\textwidth]{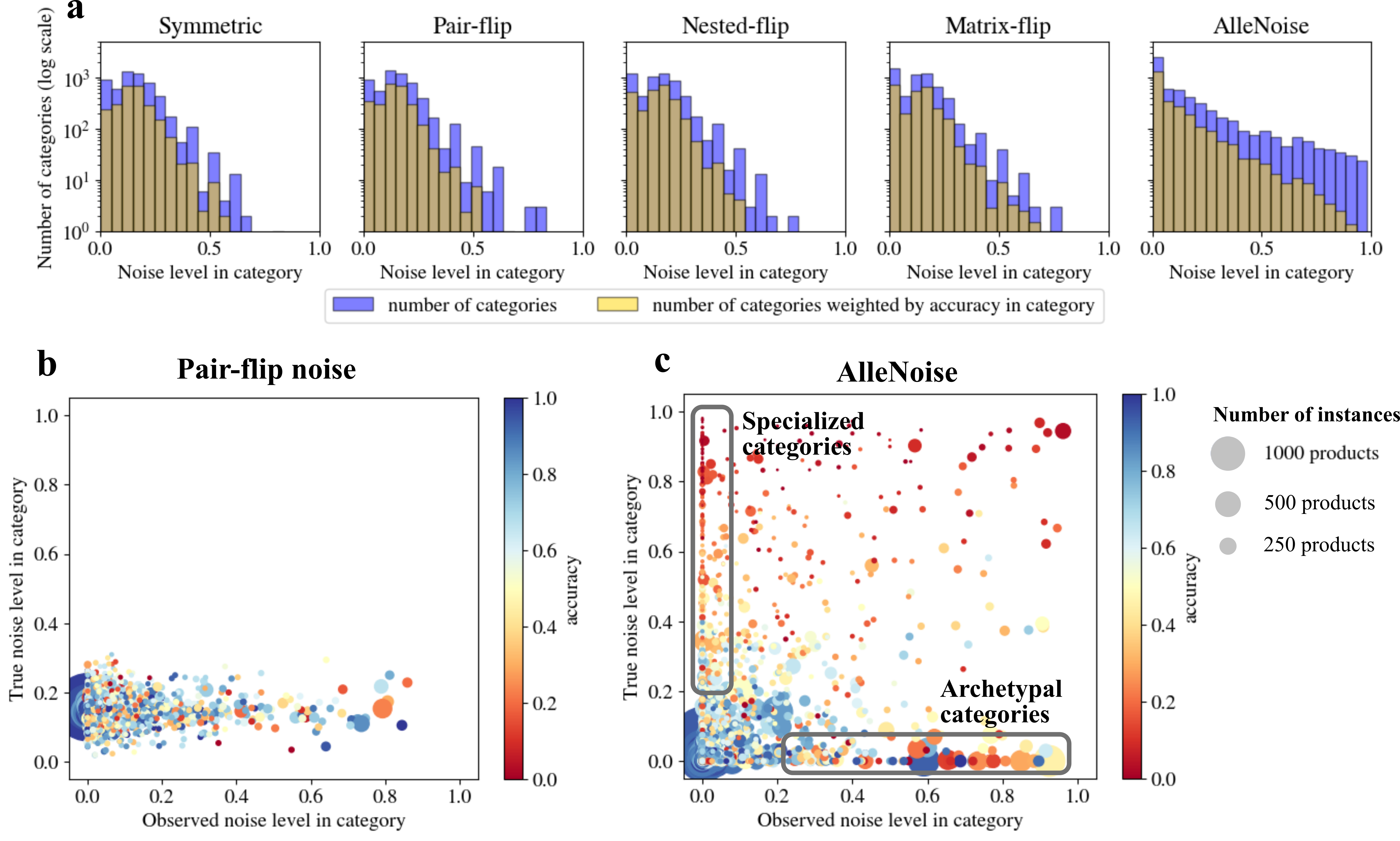}
    \caption{Noise distribution and patterns of wrong predictions across different noise types. \textbf{(a)} Noise level distribution over target categories (blue bars) shows that \textit{AlleNoise} has a substantial fraction of classes with noise level over 0.5, contrary to synthetic noise. The same distribution multiplied by per-bin macro accuracy (yellow bars) shows that those specialized categories are particularly difficult to predict correctly. \textbf{(b)} Scatter plot of true noise level versus observed noise level in each category for pair-flip noise. Marker color represents accuracy, and marker size reflects category size. True noise levels are concentrated around 15\%, with no distinct specialized or archetypal categories observed. The plot includes only categories with at least 10 products. \textbf{(c)} Scatter plot for real-world \textit{AlleNoise}, highlighting the presence of many specialized and archetypal categories. Accuracy in specialized categories is negatively correlated with the true noise level. A significant number of categories exhibit both high true noise and high observed noise levels. Scatter plots for other noise types are presented in \textbf{Fig.~\ref{fig:scatter_plots}}, Appendix~\textbf{\ref{appendix:uneven_gains}}.}
    \label{fig:noise_level}
\end{figure}

To get even more insight into why the real-world noise in \textit{AlleNoise} is more challenging than synthetic noise types, we analyzed the class distribution within our dataset. 
% It is clear that most of the noisy instances belong to a relatively small number of categories (\textbf{Fig.~\ref{fig:noise_level}}). 
For synthetic noise types, there are very few highly-corrupted categories (\textbf{Fig.~\ref{fig:noise_level}a, b}). On the other hand, 
% when we look at the class distribution 
for \textit{AlleNoise}, 
% it is evident that there is
there is a significant number of 
% highly-corrupted 
such
categories (\textbf{Fig.~\ref{fig:noise_level}a, c}).
The baseline model test accuracy is much lower for these classes than for other, less corrupted, categories.
% It is particularly difficult to correctly predict:
The set of those highly-corrupted classes is heavily populated by the following:
\begin{itemize}
    \item \textit{Specialized categories} that can be easily mistaken for a more generic category. For example, items belonging to the class \textit{safety shoes} are frequently listed in categories \textit{derby shoes}, \textit{ankle boots} or \textit{other}. In these cases, the model encounters a high true noise level, as it sees a large number of mislabeled instances with very few correctly labeled ones, which hinders its ability to learn accurate class associations (\textbf{Fig.~\ref{fig:noise_level}c}).
    % representations.
    \item 
    \textit{Archetypal categories}
    that are considered the most representative examples of a broader parent category. For instance, car tires are most frequently listed in \textit{Summer tires} even when they actually should belong to \textit{All-season tires} or other specialized categories. In this scenario, the model encounters a high observed noise level, as it sees a large number of specialized items mislabeled as the archetypal class, distorting its learned representation of that category
\end{itemize}

We hypothesize that these categories, with their respective high true and observed noise levels, are the primary contributors to the models' poor performance on \textit{AlleNoise} - an issue not present in synthetic noise types, which fail to model the complexity of specialized and archetypal categories (\textbf{Fig.~\ref{fig:noise_level}b} and \textbf{\ref{fig:scatter_plots}}). This challenge is further compounded by the inability of the benchmarked methods to consistently improve accuracy across categories with varying levels of both synthetic and real-world noise. See Appendix~\textbf{\ref{appendix:uneven_gains}} for further discussion.

\section{Discussion}
\label{section:discussion}

Our experiments show that the real-world noise present in \textit{AlleNoise} is a challenging task for existing methods for learning with noisy labels. We hypothesize that the main challenges for these methods stem from two major features of \textit{AlleNoise}: 1) real-world, instance dependent noise distribution, 2) relatively large number of categories with class imbalance and long tail. While previous works have investigated challenges 1)~\cite{wei_learning_2022} and 2)~\cite{wu_noisywikihow_2023}, this paper combines both in a single dataset and evaluation study, while also applying them to text data. 
We hope that making \textit{AlleNoise} available publicly will spark new method development, especially in directions that would address the features of our dataset.

Based on our experiments, we make several interesting observations. The methods that rely on removing examples from within a batch perform noticeably worse than other approaches. We hypothesize that this is due to the large number of classes and the unbalanced distribution of their sizes (especially the long tail of underrepresented categories) in \textit{AlleNoise} - by removing samples, we lose important information that is not recoverable. This is supported by the fact that such noise filtration methods excel on simple benchmarks like CIFAR-10, which all have a completely different class distribution. In order to mitigate the noise in \textit{AlleNoise}, a more sophisticated approach is necessary. A promising direction seems to be the one presented by ELR. While for the real-world noise it did not increase the results above the baseline CE, it was the best algorithm for class-dependent noise types.
The outstanding performance of ELR might be attributed to its target smoothing approach. The use of such soft labels may be particularly adequate to extreme classification scenarios where some of the classes are semantically close. Extending this idea to include an instance-dependent component may lead to an algorithm robust to the real-world noise in \textit{AlleNoise}. Furthermore, based on the results of the memorization metric, it is evident that this realistic noise pattern needs to be tackled in a different way than synthetic noise. With the clean labels published as a part of \textit{AlleNoise}, we enable researchers to further explore the issue of memorization in the presence of real-world instance-dependent noise.

\section{Limitations}
\label{section:limitations}
Our dataset presents several notable characteristics and limitations. It includes $\sim$500,000 product titles from over 5,000 categories, sampled to reflect the broader product catalog while controlling for the tangible 15\% noise level. However, the dataset's focus on a Polish marketplace might limit its diversity and applicability to other regions, particularly outside the EU. The specialized nature of e-commerce text might not be completely transferable to other NLP domains. Moreover, the translation accuracy of our in-house scale neural machine translation system remains imperfect, which has an impact on classification accuracy (see Appendix~\textbf{\ref{appendix-translation}}). Despite these challenges, the \textit{AlleNoise} dataset is a useful resource for benchmarking text classification models, especially with its known noise level, distinguishing it from other e-commerce datasets \cite{hou_bridging_2024,lin_dataset_2019,akritidis_self-verifying_2020, akritidis_effective_2018}. For an extended discussion on limitations of AlleNoise please refer to Appendix~\textbf{\ref{appendix:limitations}}.

\section{Conclusions and future work}

In this paper, we presented a new dataset for the evaluation of methods for learning with noisy labels. Our dataset, \textit{AlleNoise}, contains a real-world instance-dependent noise distribution, with both clean and noisy labels. It provides a large-scale classification problem, and unlike most previously available datasets in the field of learning from noisy labels, features textual data. We performed an evaluation of established noise-mitigation methods, which showed quantitatively that these approaches are not enough to alleviate the noise in our dataset. With \textit{AlleNoise}, we hope to jump-start the development of new robust classifiers that would be able to handle demanding, real-world instance-dependent noise, reducing errors in practical applications of text classifiers.

The scope of this paper is limited to BERT-based classifiers. As \textit{AlleNoise} includes clean label names in addition to noisy labels, it could be used to benchmark Large Language Models in few-shot or in-context learning scenarios. We leave this as a future research direction.

\begin{ack}
\subsection*{Acknowledgements}
We thank Mikołaj Koszowski for his help with translating the product titles. We also thank Karol Jędrzejewski for his help in pin-pointing the location of appropriate data records in Allegro data warehouse.

\subsection*{Funding}
This work was funded fully by Allegro.com.

\subsection*{Competing interests}
We declare no competing interests.
\end{ack}

\bibliographystyle{abbrv}

\newpage
\appendix

\section{Implementation details}
\label{appendix:impl_details}

\subsection*{Self-Paced Learning} 

The Self-Paced Learning (SPL)~\cite{kumar_self-paced_2010} method sets a threshold $\lambda$ value for the loss and all examples with loss larger than $\lambda$ are skipped, since they are treated as hard to learn (because they are possibly noisy). After each training epoch, the threshold is increased by some constant multiplier. For simplification, we adjusted SPL in the following manner.

We set a parameter $\tau_{SPL}$, which controls the percentage of samples with the highest loss within a batch that are excluded. The value of $\tau_{SPL}$ should be equal to the noise level present in the training dataset. As such, at each step, we exclude a set percentage of potentially noisy examples, thus reducing the impact of label noise on the training process. We keep the value of $\tau_{SPL}$ constant throughout the training.

\subsection*{Provably Robust Learning}

The Provably Robust Learning (PRL)~\cite{liu_learning_2021} algorithm works in a similar manner to SPL. We follow the authors by introducing the $\tau_{PRL}$ parameter, which controls the percentage of samples excluded from each training batch on the basis of their gradient norm. Specifically, $\tau_{PRL}\%$ of samples with highest gradient norm are omitted, while the rest is used to update model parameters. The value of $\tau_{PRL}$ should be equal to the noise level in the training dataset.

\subsection*{Clipped Cross-Entropy}

Since our implementation of SPL doesn't have a hard loss threshold, we introduce a simple Clipped Cross-Entropy (CCE) baseline to check the effectiveness of such an approach. The CCE method checks if the loss is greater than some threshold $\lambda_{CCE}$. If so, the loss is clipped to that value. Otherwise, it is left unchanged. Thus, we always use all training samples, but the impact of label noise is alleviated by clipping the loss.

\subsection*{Early Learning Regularization} 

For Early Learning Regularization (ELR)~\cite{liu_early-learning_2020}, we followed the implementation published by the authors. 
We compute the softmax probabilities for each sample in a batch and clamp them, then compute the soft targets via temporal ensembling and use these targets in the loss function calculation.
Our implementation includes one step not present in the publication text - softmax probability clamping in range $[\epsilon, 1 - \epsilon]$, where $\epsilon$ is a clamp margin parameter. Aside from this, we use the $\beta$ target momentum and $\lambda_{ELR}$ regularization parameters just as they were presented by the authors.

\subsection*{Generalized Jensen-Shannon Divergence Loss}

The Generalized Jensen-Shannon Divergence (GJSD)~\cite{englesson_generalized_2021} loss function is a generalization of Cross-Entropy (CE) and Mean Absolute Error (MAE) losses. We follow the implementation provided by the authors, in which we use the $M$ parameter to set the number of averaged distributions and the $\pi$ parameter to adjust the weight between CE and MAE. While the authors share separate implementations for GJSD with and without consistency regularization, we implement it as a toggle to make the code more uniform. Since consistency regularization requires data augmentation and the GJSD authors described only augmentations for the image domain, we implemented several textual augmentations of our own: random token dropping, consecutive token dropping, random token swapping. However, in our experiments, we have kept consistency regularization turned off due to its detrimental effect on model convergence and test accuracy.

\subsection*{Co-teaching}

While the methods described above modified the loss function in various ways, Co-teaching (CT)~\cite{han_co-teaching_2018} works in a different manner. It requires optimizing two sets of model parameters at the same time. As such, following the algorithm described by the authors, we implemented a custom model class, which manages the update of these two sets of weights and the exchange of low-loss examples at each optimization step. We keep the parameters $k$ and $\tau_{CT}$, to control the starting epoch for CT and the noise level (i.e. the percentage of low-loss examples that are exchanged between networks), respectively.

\subsection*{Co-teaching+}

For Co-teaching+ (CT+)~\cite{yu_how_2019}, we again adhere to the algorithm described by the authors. We use the same implementation framework as for CT, adjusting only the sample selection mechanism to look within examples for which there is disagreement between the two networks. Following the advice in the publication text, we use the \textit{recommended} update strategy for the fraction of instances to select, which is calculated based on the epoch number, as well as parameters $k$ and $\tau_{CT+}$.

\subsection*{Mixup}

The Mixup (MU)~\cite{zhang_mixup_2018} technique keeps the loss function (CE) and the hyperparameters of the baseline model unchanged, only augmenting the training data during the training procedure. We use in-batch augmentation, fixed per-batch mixing magnitude sampled from $Beta(\alpha, \alpha)$ (where $\alpha$ is provided as input), and the mixed pairs are sampled without replacement from that distribution. Since we cannot mix input in the same way as for images, we implemented in-batch augmentation for logits. In addition, we also keep the $r_{MU}$ ratio parameter, to adjust the percentage of the batch size which is taken for augmentation in MU. Note: our hyperparameter tuning procedure resulted in setting both $\alpha$ and $r_{MU}$ to low values (\textbf{Tab.~\ref{tab:hyperparams}}), contrary to what is recommended by the authors.

\subsection*{Pseudo-Label Selection}

For the Pseudo-Label Selection~\cite{albert_noisecorrection_pls_2022} algorithm, we use the code provided by the authors. The method relies on image augmentations during training. To adapt these transformations to text modality, we utilize the \textit{nlpaug}~\cite{ma_nlpaug_2019} Python package. For weak augmentations, we use random word swaps with probability~0.3. For strong augmentation, we again apply random word swaps, followed by random word crops and random word splits, all with probability~0.3. We tune two hyperparameters: the number of epochs that are trained with standard cross-entropy $k_{warm}$, and the size of the contrastive projection layer $h_{proj}$.

\setcounter{table}{0}
\setcounter{figure}{0}
\renewcommand{\thetable}{S\arabic{table}}
\renewcommand{\thefigure}{S\arabic{figure}}

\begin{table}[h]
    \centering
    \captionsetup{labelfont=bf}
    \begin{tabular}{ccc}
    \toprule
    Method & Hyperparameters & Selected values \\
    \midrule
    SPL & $\tau_{SPL}$ & equal to noise level \\
    PRL & $\tau_{PRL}$ & equal to noise level \\
    ELR & $\epsilon, \beta, \lambda_{ELR}$ & 1e-5, 0.6, 2 \\
    CCE & $\lambda_{CCE}$ & 9.5 \\
    MU & $\alpha, r_{MU}$ & 0.1, 0.1 \\
    GJSD & $M, \pi$ & 2, 0.001 \\
    CT & $k$, $\tau_{CT}$ & 8, equal to noise level \\
    CT+ & $k$, $\tau_{CT+}$  & 8, equal to noise level \\ 
    PLS & $k_{warm}$, $h_{proj}$ & 8, 512 \\
    \bottomrule
    \end{tabular}
    \caption{Hyperparameter values for all benchmarked methods, selected through a tuning procedure.}
    \label{tab:hyperparams}
\end{table}

\section{Results of experiments with higher noise level}
\label{appendix:higher_noise}

% To differentiate the performance of evaluated methods on synthetic data, we perform another series of experiments. We sub-sample clean examples, and create a reduced clean dataset. Then we create noisy datasets containing the same number of noisy examples as before, resulting in datasets with 40\% of noisy observations. 
% Results in Table~\ref{tab:pretrained-noise40} show similar pattern as in  Table~\ref{tab:pretrained-noise15} - evaluated methods improve the results compared to baseline CE on synthetic data, but fail to do so on \textit{AlleNoise}.
% However, higher level of noise have let us differentiate between types of synthetic noise. We clearly observe that asymmetric noise is harder than symmetric, and that asymmetric noises designed to mimic real-world noise  closer are harder. However, even on asymmetric matrixflip noise, we do not get as low accuracy as on \textit{AlleNoise}.

For completeness, we evaluate the accuracy for all methods on datasets with 40\% synthetic noise (\textbf{Tab.~\ref{tab:pretrained-noise40}}). The best methods for this noise level are the same as for the case of 15\% noise: for symmetric noise, GJSD is the best method, while for asymmetric noise types it is ELR. However, it is clear that some methods show more noticeable effect when compared to the baseline for the 40\% noise level than for the 15\%. While MU and CCU stay close to the baseline results for all noise types and SPL underperforms in all cases, CT consistently gives an improvement over the baseline and CT+ decreases the result for the symmetric noise, but is better than the baseline for asymmetric noise types.

\begin{figure}[h]
    \captionsetup{labelfont=bf}
    \centering
    \includegraphics[width=\textwidth]{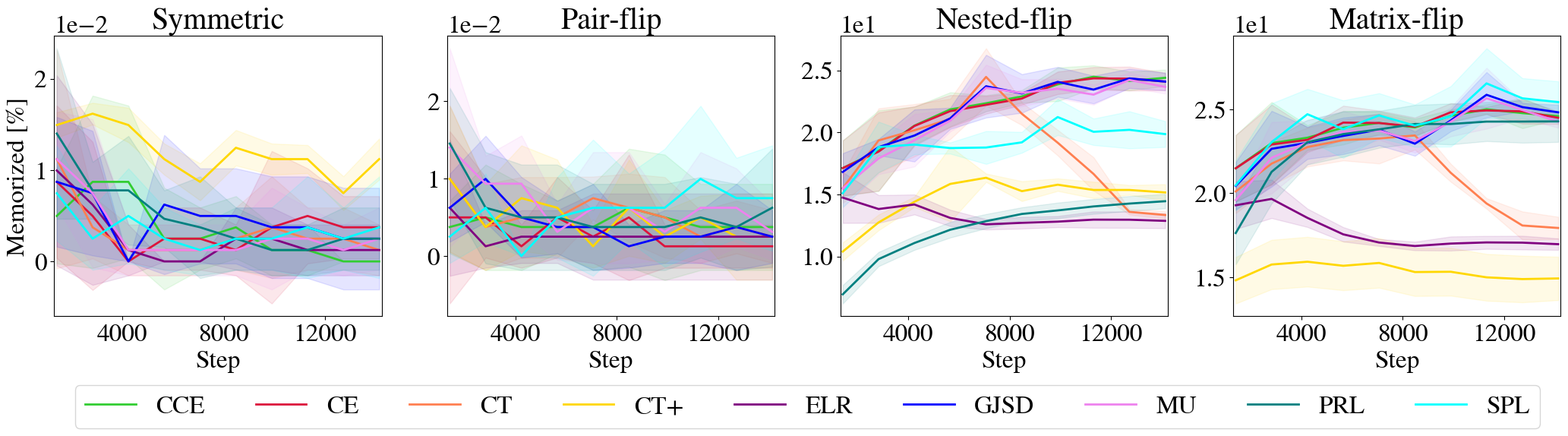}
    \caption{Value of $\texttt{memorized}_{val}$ for different noise types, measured at each training step. In all cases the noise level was set at 40\%.}
    \label{fig:memorization-40}
\end{figure}

We also plot $\texttt{memorized}_{val}^{\texttt{noisy}}$ for those datasets (\textbf{Fig.~\ref{fig:memorization-40}}). For symmetric and pair-flip noise types the memorization for all methods is very low. For nested-flip and matrix-flip it is a bit higher, indicating that these two noise types are more challenging, and thus induce more memorization in the model.

\vspace{20pt}

\vspace{20pt}

\begin{table}[h]
\centering
\captionsetup{labelfont=bf}
\begin{tabular}{lr|rrrr}
\toprule
% \multicolumn{1}{c}{} & \multicolumn{1}{c}{Symmetric} & \multicolumn{1}{c}{\begin{tabular}[c]{@{}c@{}}Asymmetric \\ pair-flip\end{tabular}} & \multicolumn{1}{c}{\begin{tabular}[c]{@{}c@{}}Asymmetric \\ nested-flip\end{tabular}} & \multicolumn{1}{c}{\begin{tabular}[c]{@{}c@{}}Asymmetric \\ matrix-flip\end{tabular}} \\
& \multicolumn{1}{c|}{Clean set}            & \multicolumn{1}{c}{Symmetric}   & \multicolumn{1}{c}{Pair-flip}   & \multicolumn{1}{c}{Nested-flip} & \multicolumn{1}{c}{Matrix-flip} \\
\midrule
CE & \textbf{74.85 ± 0.15} & 67.29 ± 0.12 & 55.18 ± 0.26 & 52.87 ± 0.19 & 54.04 ± 0.23 \\ \midrule
ELR & 74.81 ± 0.11 & 67.23 ± 0.18 & \textbf{66.12 ± 0.15} & \textbf{62.27 ± 0.19} & \textbf{61.72 ± 0.23} \\ %\midrule
MU & 74.73 ± 0.09 & 67.14 ± 0.14 & 55.28 ± 0.26 & 52.38 ± 0.24 & 54.26 ± 0.25 \\ %\midrule
CCE & 74.80 ± 0.09 & 68.92 ± 0.14 & 55.13 ± 0.28 & 52.07 ± 0.58 & 54.02 ± 0.18 \\ %\midrule
CT & *74.85 ± 0.15 & 68.60 ± 0.14 & 60.49 ± 0.24 & 58.48 ± 0.28 & 57.69 ± 0.47 \\ %$ \midrule
CT+ & *74.85 ± 0.15 & $\downarrow$64.67 ± 0.32 & 59.03 ± 0.42 & 56.16 ± 0.42 & 57.06 ± 0.29 \\
PRL & *74.85 ± 0.15 & $\downarrow$65.01 ± 0.30 & 62.22 ± 0.45 & 56.39 ± 0.40 & 51.59 ± 0.97 \\% \midrule
SPL & *74.85 ± 0.15 & 65.27 ± 0.35 & $\downarrow$44.92 ± 1.52 & $\downarrow$42.29 ± 1.06 & $\downarrow$40.89 ± 0.70 \\% \midrule
GJSD & 74.63 ± 0.10 & \textbf{69.80 ± 0.12} & 54.78 ± 0.30 & 51.92 ± 0.46 & 53.84 ± 0.10 \\ 
\bottomrule
\end{tabular}
\caption{Accuracy of the evaluated methods on the clean dataset compared to various noisy datasets with 40\% noise level. The noisy datasets include symmetric synthetic noise and asymmetric synthetic noise types: pair-flip, nested-flip, and matrix-flip. * marks cases equivalent to the baseline CE. $\downarrow$ marks results significantly worse than the baseline CE. Best results for each noise type are bolded.}
\label{tab:pretrained-noise40}
\end{table}

\section{Training, validation and test accuracy of baseline CE}
\label{appendix:accuracy}

\begin{figure}[h]
    \captionsetup{labelfont=bf}
    \centering
    \includegraphics[width=1\textwidth]{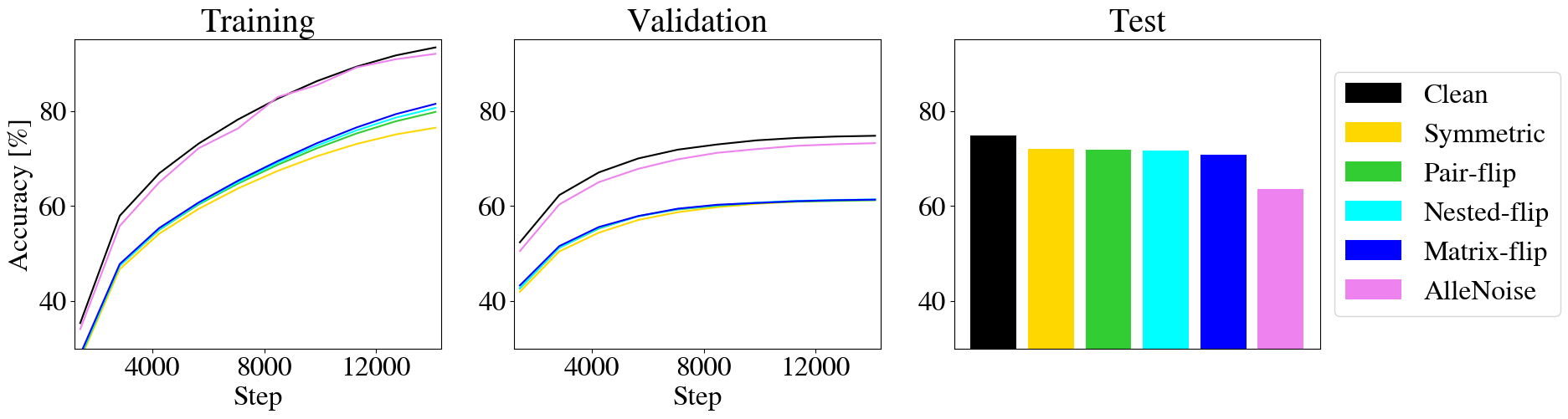}
    \caption{Accuracy of CE for different 15\% noise types and the clean dataset. Training and validation accuracy is computed at every validation step during training, test accuracy is computed only for the final model.}
    \label{fig:accuracy-15}
\end{figure}

We measure training, validation and test accuracy of the baseline CE method on clean and noisy datasets (\textbf{Fig.~\ref{fig:accuracy-15}}). Training accuracy and validation accuracy are measured with observed (potentially noisy) labels, while test accuracy is measured with hidden, clean labels. The real-world noise in \textit{AlleNoise} differs from synthetic noise types. When it comes to training accuracy, the close proximity of \textit{AlleNoise} and clean training curves indicates that the real-world noise distribution is of similar nature to the distribution of clean labels. In other words, the decision boundary for AlleNoise is easier to fit than for synthetic noise types. 

For the validation accuracy, we can observe that the memorized knowledge from \textit{AlleNoise} transfers to validation instances with noisy labels, as indicated by the small distance between the validation accuracy curves for \textit{AlleNoise} one for the clean dataset. This is not the case for synthetic noise types. This shows that \textit{AlleNoise} is indeed like real, clean training data - the model learns from it and generalizes from it just like from clean data. Synthetic noise types do not have this property.

Test accuracy on \textit{AlleNoise} is the lowest of all considered noise types, which shows that the model is misled by the mislabeled training data. Synthetic noisy labels do not mislead the model to such an extent, as even the baseline CE managed to ignore a certain amount of those errors, leading to higher test accuracy than on \textit{AlleNoise}.

% \begin{figure}[h]
%     \captionsetup{labelfont=bf}
%     \centering
%     \includegraphics[width=\textwidth]{figures-results/final/accuracies.png}
%     \caption{\textcolor{blue}{Accuracy of CE for different noise types and clean dataset, measured at each training step. In all cases the noise level was set at 15\%. Dashed lines denote training accuracy, solid lines denote validation accuracy. Test accuracy is marked with a dot.}}
%     \label{fig:accuracy-15}
% \end{figure}

% \begin{landscape}
% \end{landscape}

\section{Uneven gains across noise levels \label{appendix:uneven_gains}
and types}
\begin{figure}[h]
    \captionsetup{labelfont=bf}
    \centering
    \includegraphics[width=\textwidth]{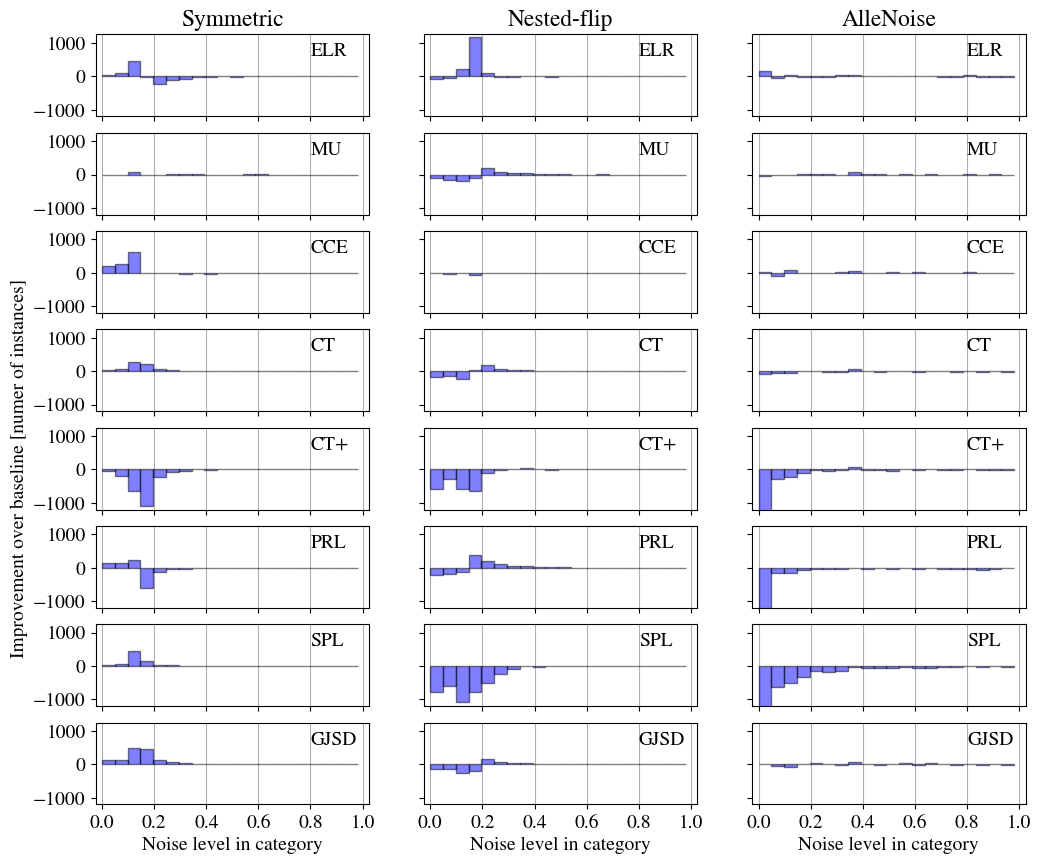}
    \caption{Histograms illustrating the performance of eight learning methods across different noise levels for three noise types (symmetric, nested-flip, and \textit{AlleNoise}). The x-axis represents the noise level within each category, while the y-axis shows the difference in the number of correctly predicted instances between each method and the baseline. Positive bars indicate accuracy gains, while negative bars indicate accuracy losses. The sum of all bars corresponds to the total gain in correctly predicted instances, which determines the final accuracy when normalized by the total number of instances in the dataset.}
    \label{fig:improvement}
\end{figure}

Learning from real-world noisy labels in imbalanced datasets like \textit{AlleNoise} is particularly challenging, as it demands consistent accuracy improvements across varying noise levels (\textbf{Fig.~\ref{fig:noise_level}}). To assess the performance of benchmarked methods under diverse noise conditions, we analyzed accuracy gains across different noise levels and types (\textbf{Fig.~\ref{fig:improvement}}). Our findings reveal the following:
\begin{itemize}
\item All methods except CT+ handled symmetric noise effectively up to 15\%, with GJSD providing consistent improvements across the entire noise range.
\item ELR’s poor performance in categories with symmetric noise above 20\% neutralized its gains at noise levels below 15\%.
\item Most methods (ELR, MU, CT, PRL, GJSD) performed well with asymmetric noise between 15\% and 25\% but struggled with noise below 10\%.
\item Noise-filtering methods (PRL, SPL, CT, CT+) often undermined the model's performance in categories with asymmetric noise below 15\%, likely due to excessive instance removal.
\item Gains by MU, CT, CT+, PRL, and GJSD on asymmetric noise between 25\% and 40\% were balanced out by their poor performance in categories with noise below 10\%.
\end{itemize}

In summary, none of the methods successfully modeled both high and low levels of synthetic noise. This limitation is even more pronounced in the real-world \textit{AlleNoise} dataset, where the majority of instances (56\%) are in categories with very low noise levels (< 5\%), and there is a long tail (11\%) of instances in highly corrupted classes (noise levels > 40\%). These characteristics of the real-world noise distribution make the \textit{AlleNoise} dataset a challenge for classification models, setting it apart from both synthetic noise scenarios and other benchmarks in the field of learning with noisy labels.

\begin{figure}[h]
    \captionsetup{labelfont=bf}
    \centering
    \includegraphics[width=\textwidth]{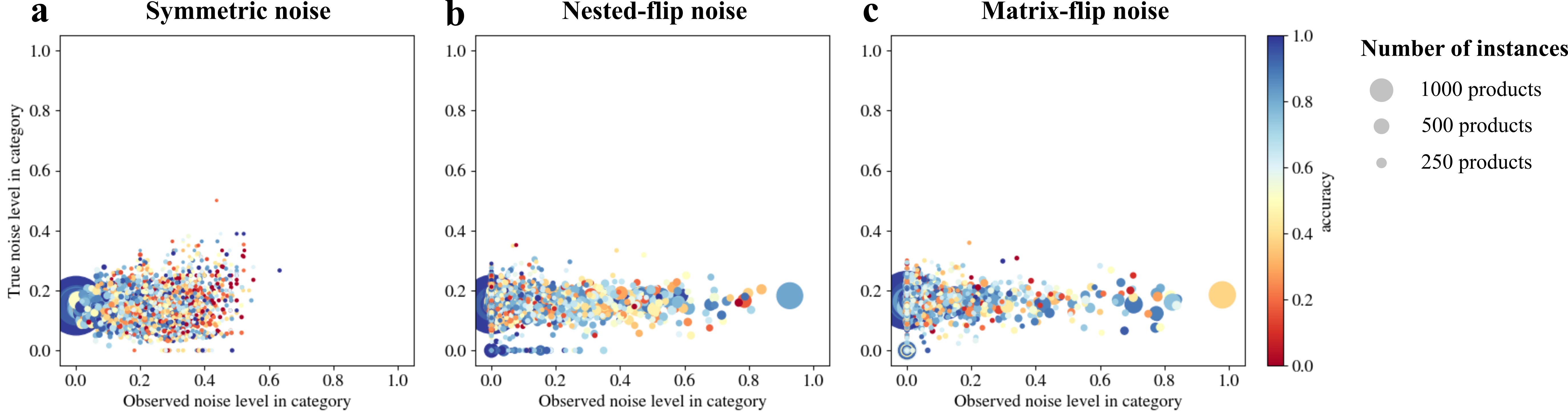}
    \caption{Noise distribution and patterns of wrong predictions for \textbf{(a)} symmetric, \textbf{(b)} nested-flip, and \textbf{(c)} matrix-flip noise types. Scatter plots depict true noise level versus observed noise level in each category, with marker color representing accuracy and diameter proportional to category size. All noise types show true noise concentrated around 15\%. Symmetric noise fails to capture categories with high observed noise levels, while nested-flip noise models such categories better and also includes many noise-free categories, similar to real-world noise. Matrix-flip noise successfully captures archetypal categories with low true noise and high observed noise. However, all types struggle to model specialized categories.}
    \label{fig:scatter_plots}
\end{figure}

% To differentiate the performance of evaluated methods on synthetic data, we perform another series of experiments. We sub-sample clean examples, and create a reduced clean dataset. Then we create noisy datasets containing the same number of noisy examples as before, resulting in datasets with 40\% of noisy observations. 
% Results in Table~\ref{tab:pretrained-noise40} show similar pattern as in  Table~\ref{tab:pretrained-noise15} - evaluated methods improve the results compared to baseline CE on synthetic data, but fail to do so on \textit{AlleNoise}.
% However, higher level of noise have let us differentiate between types of synthetic noise. We clearly observe that asymmetric noise is harder than symmetric, and that asymmetric noises designed to mimic real-world noise  closer are harder. However, even on asymmetric matrixflip noise, we do not get as low accuracy as on \textit{AlleNoise}.

\clearpage

\section{Exploratory Data Analysis}
\label{appendix:eda}

To facilitate the data preprocessing and feature selection stages of future machine learning model development, we provide a detailed analysis of the data in the \textit{AlleNoise} dataset.

\subsection{Quality of product titles}
During the data sampling stage, we removed instances with titles that hasn't met several quality standards:
\begin{itemize}
\setlength\itemsep{0em}
    \item were too short,
    \item were duplicated,
    \item were highly repetitive,
    \item included errors from translation services,    
    \item contained only digits,    
    \item contained derogatory words.
\end{itemize}
There are no conflicting instances with the same title and different labels, even when lowercase texts are considered. The ratio of special characters (non-alphanumerical characters) is 2\%.

\begin{figure}[ht]
    \captionsetup{labelfont=bf}
    \centering
    \includegraphics[width=0.8\textwidth]{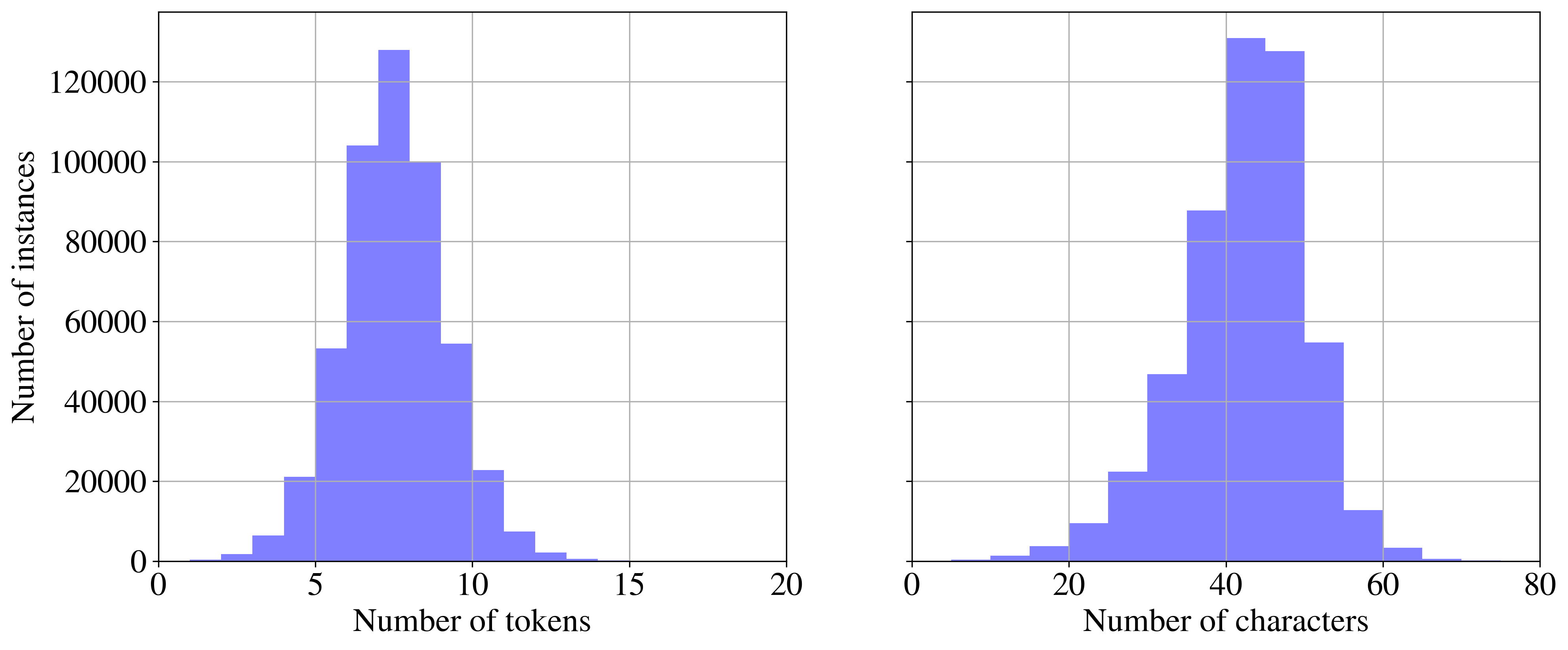}
    \caption{Distribution of product title length in \textit{AlleNoise}. The number of tokens corresponds to the number of words in the product title.}
    \label{fig:eda-text-length}
\end{figure}

The median number of words is 7 (percentiles: $5^{\text{th}} = 4, 95^{\text{th}} = 10$), while the median number of characters is 43 ($5^{\text{th}} = 27, 95^{\text{th}} = 53$) (\textbf{Fig.~\ref{fig:eda-text-length}}). 

\subsection{Semantic contents of the dataset}
\begin{figure}[ht]
    \captionsetup{labelfont=bf}
    \centering
    \includegraphics[width=\textwidth]{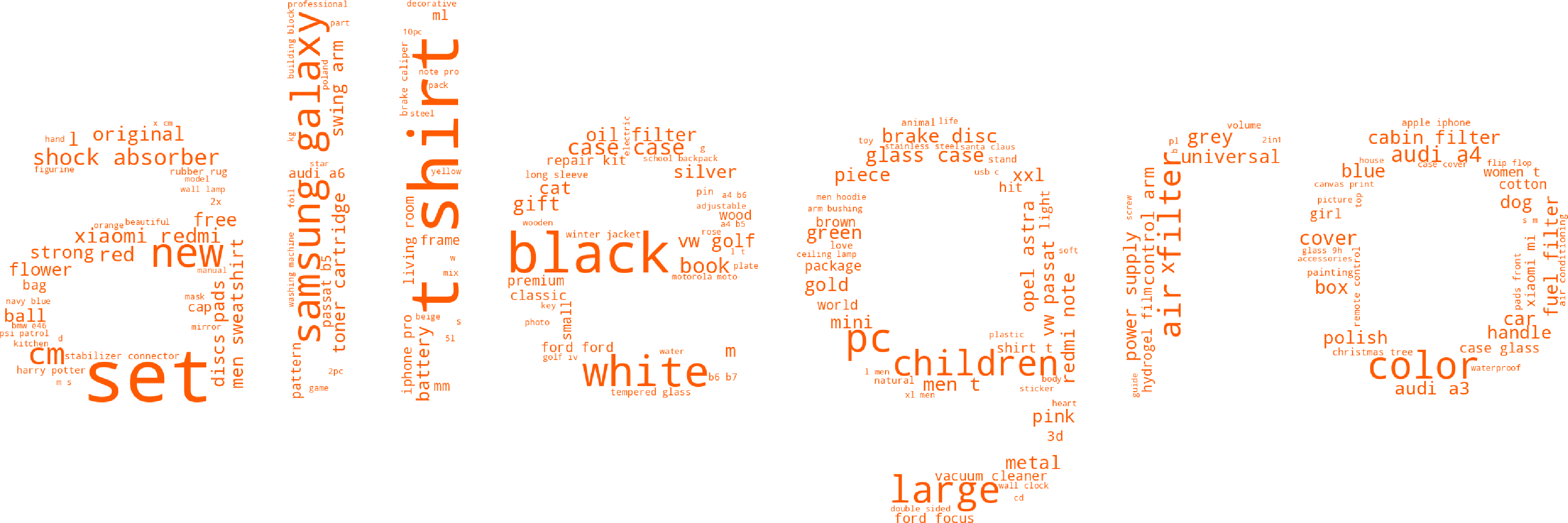}
    \caption{Wordcloud with top-200 n-grams from \textit{AlleNoise} dataset.}
    \label{fig:eda-wordcloud}
\end{figure}

\begin{figure}[ht]
    \captionsetup{labelfont=bf}
    \centering
    \includegraphics[width=0.8\textwidth]{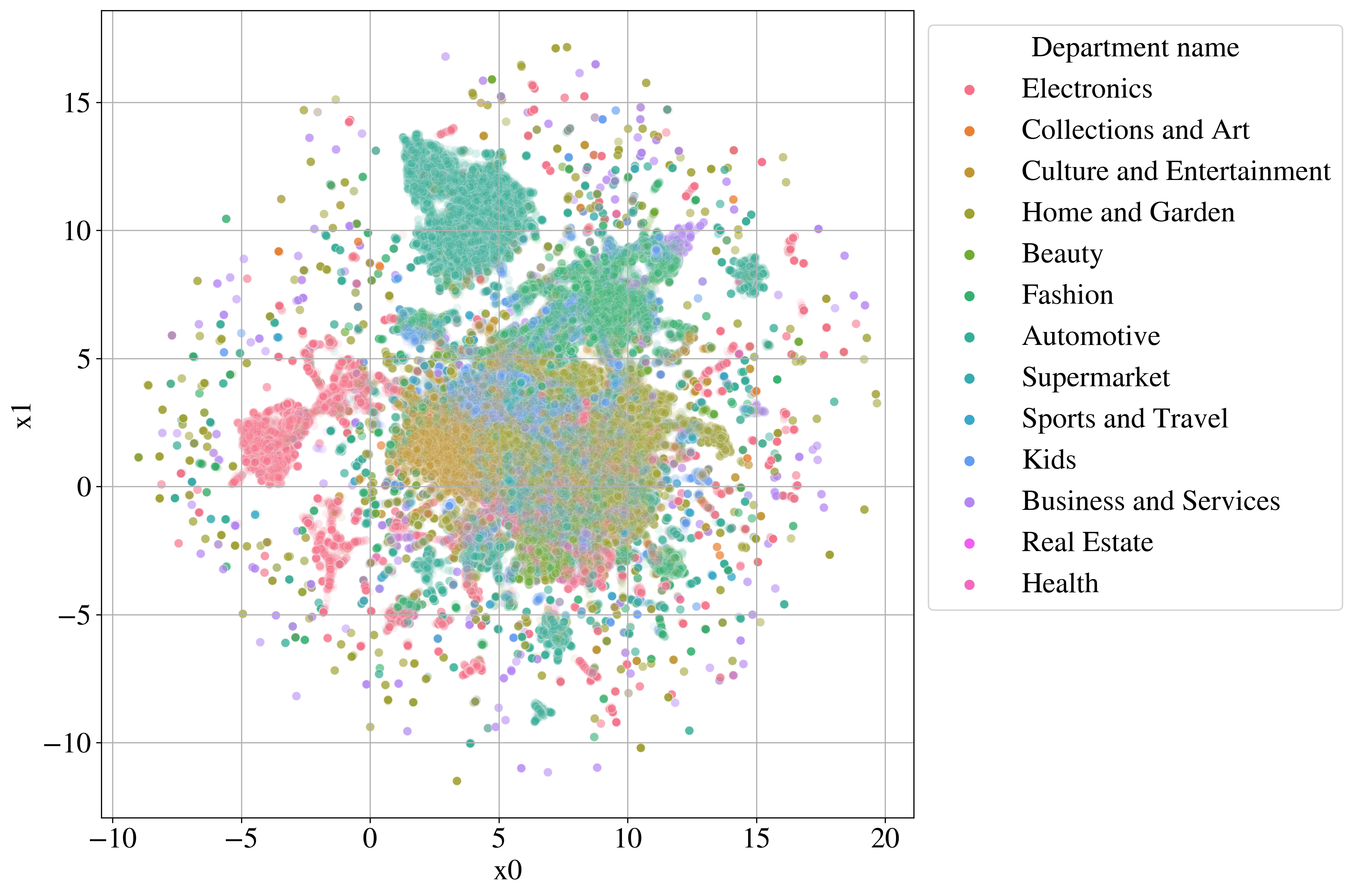}
    \caption{Visualization of the title embedding space. A pretrained USE-v4 model was used as the text embedder. Titles were lowercased before vectorization. A 2D visualization was generated via UMAP dimensionality reduction (number of neighbors: 15, metric: euclidean). Points are colored by department name (meta-category).}
    \label{fig:eda-use-umap}
\end{figure}

The top-5 popular n-grams in the dataset are: ''set'', ''t shirt'', ''black'', ''new'' and ''white'' (\textbf{Fig.~\ref{fig:eda-wordcloud}}). Product names are not easily separable in the embedding space, therefore, the classification task should be considered challenging (\textbf{Fig.~\ref{fig:eda-use-umap}}).

\subsection{Label distribution}
Dataset sampling followed the distribution of products on Allegro, hence the dataset closely mimics the assortment diversity and the proportion of popular to unpopular categories. The ''Home and Garden'' department has the highest number of categories (994), while ''Automotive'' has the most instances (106374) (\textbf{Tab.~\ref{tab:eda-department-all}}). ''Home and Garden'', ''Business and Services'' and ''Electronics'' lead in terms of the amount of noisy labels (\textbf{Tab.~\ref{tab:eda-department-transfers}}), when the true underlying category is considered. 17\% of mislabeled categories were corrected to a category from a department different than the initial one.

% All categories:
\begin{table}[ht]
    \centering
    \captionsetup{labelfont=bf}
    \begin{tabular}{lll}
    \toprule
    Department name           & \# categories & \# instances \\
    \midrule
    Automotive                & 757 & $\boldsymbol{106374}$        \\
    Home and Garden           & $\boldsymbol{994}$ & 96781         \\
    Fashion                   & 223 & 66839         \\
    Electronics               & 837 & 60912         \\
    Culture and Entertainment & 438  & 58482         \\
    Kids                      & 604  & 39221         \\
    Business and Services     & 364 & 23089         \\
    Sports and Travel         & 436 & 13257         \\
    Supermarket               & 375 & 12669         \\
    Beauty                    & 210 & 12464         \\
    Collections and Art       & 279 & 7158          \\
    Health                    & 174 & 5058          \\
    Real Estate               & 1 & 6             \\
    \bottomrule
    \end{tabular}
    \caption{Distribution of instances and unique categories across departments in the \textit{AlleNoise} dataset.}
    \label{tab:eda-department-all}
\end{table}

% Category transfers:
\begin{table}[ht]
    \centering
    \captionsetup{labelfont=bf}
    \begin{tabular}{ll}
    \toprule
    Department name           & \# instances \\
    \midrule
    Home and Garden           & 18623         \\
    Business and Services     & 11934         \\
    Electronics               & 9236          \\
    Culture and Entertainment & 9150          \\
    Automotive                & 7326          \\
    Kids                      & 6273          \\
    Supermarket               & 2716          \\
    Collections and Art       & 2563          \\
    Sports and Travel         & 1827          \\
    Fashion                   & 1734          \\
    Beauty                    & 1610          \\
    Health                    & 1102          \\
    \bottomrule
    \end{tabular}
    \caption{Distribution of mislabeled instances across departments in the \textit{AlleNoise} dataset.}
    \label{tab:eda-department-transfers}
\end{table}

\begin{figure}[ht]
    \captionsetup{labelfont=bf}
    \centering
    \includegraphics[width=0.75\textwidth]{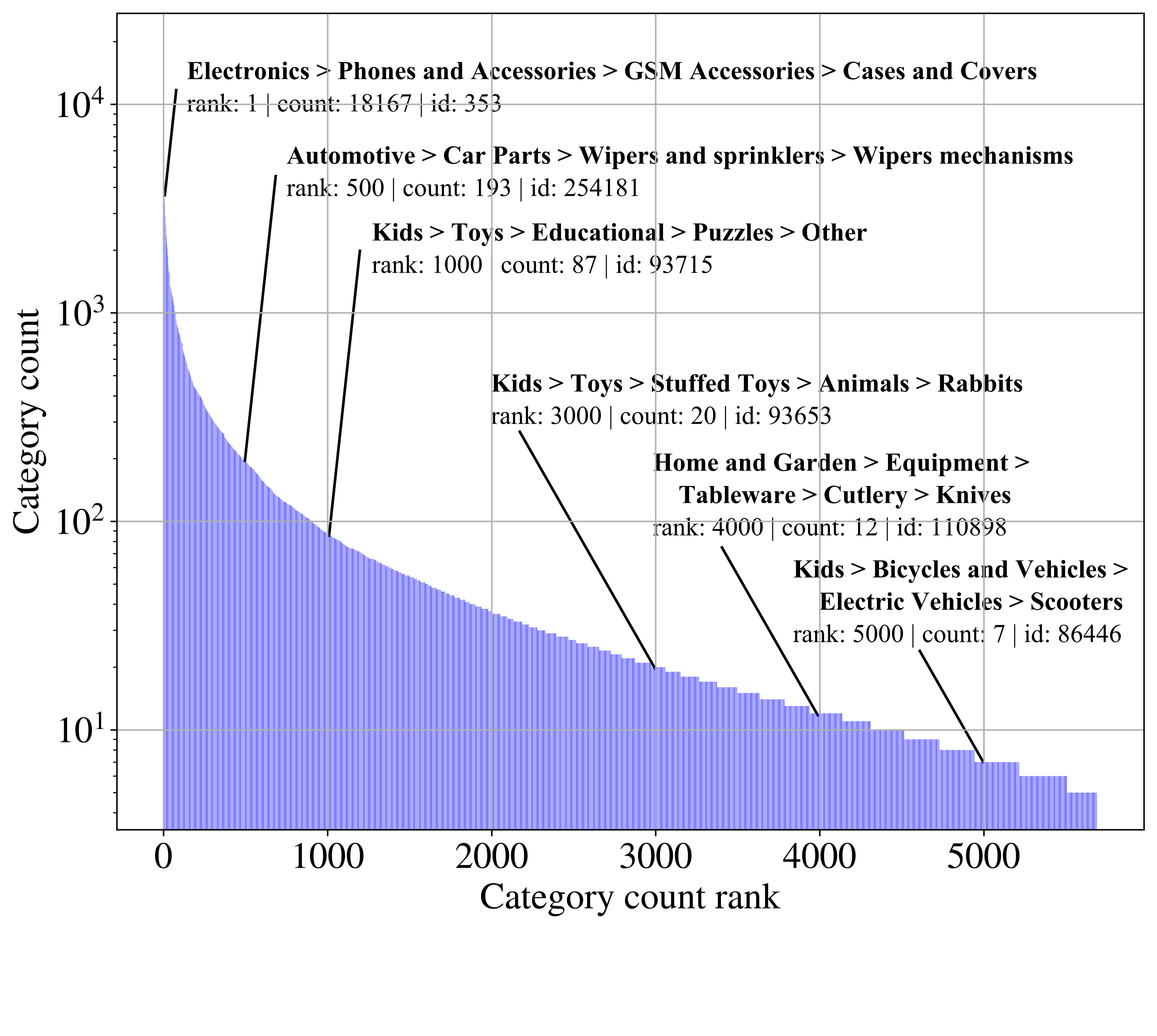}
    \caption{Distribution of category counts in the \textit{AlleNoise} dataset (log y scale). Categories are sorted by their count (x scale).}
    \label{fig:eda-longtail}
\end{figure}

While the distribution of categories in the dataset follows the true distribution of products sold in the Allegro marketplace, it poses a challenge due to its highly skewed nature (\textbf{Fig.~\ref{fig:eda-longtail}}). General, non-specialized, e-commerce platforms typically suffer from a prominent long tail, which presents significant challenges for the automated classification of products into categories. The most populated categories include phone cases, everyday clothing, and home decorations (\textbf{Tab.~\ref{tab:eda-top-categories}}). The least populated categories contain expert tools, smartphone and car models, as well as specific everyday objects (\textbf{Tab.~\ref{tab:eda-bottom-categories}}). 

% Most popular categories
\begin{table}[ht]
    \centering
    \captionsetup{labelfont=bf}
    \begin{tabular}{ll}
    \toprule
    Category name           & \# instances \\
    \midrule
    Electronics > Phones and Accessories > GSM Accessories > Cases and Covers       & 18167        \\
    Fashion > Clothing, Shoes, Accessories > Men's Clothing > T-shirts              & 8208         \\
    Home and Garden > Equipment > Decorations and Ornaments > Images and Paintings  & 6143         \\
    Fashion > Clothing, Shoes, Accessories > Men's Clothing > Hoodies               & 5984         \\
    Automotive > Car Parts > Brake System > Disc Brakes > Brake Discs               & 4122         \\
    \bottomrule
    \end{tabular}
    \caption{Top-5 categories in terms of the number of instances representing the head of the category tree.}
    \label{tab:eda-top-categories}
\end{table}

% Least popular categories
\begin{table}[ht]
    \centering
    \captionsetup{labelfont=bf}
    \begin{tabular}{ll}
    \toprule
    Category name           & \# instances \\
    \midrule
    Automotive > Cars > Passenger Cars > Tesla > Model S                  & 5        \\
    Home and Garden > Tools > Welders > Transformer Welders          & 5         \\
    Electronics > Phones and Accessories > Smartphones and Cell Phones > OnePlus > 8T               & 5         \\
    Home and Garden > Garden > Grilling > Concrete Grills   & 5         \\
    Kids > Toys > Babies > Shushers and Soothers > Other            & 5         \\
    \bottomrule
    \end{tabular}
    \caption{Bottom-5 categories in terms of the number of instances representing the long tail of the category tree.}
    \label{tab:eda-bottom-categories}
\end{table}

\section{Extended discussion of limitations} \label{appendix:limitations}

\textbf{Selection of Categories} Allegro is a general marketplace that represents a wide spectrum of products from various categories and shopping intents. Our dataset comprises over 5,000 categories sampled from nearly 23,000 overall, following the distribution of the Allegro catalog. We undersampled the entire catalog to maintain a manageable dataset size and to control noise levels at around 15\%.

\textbf{Allegro as a Polish Marketplace} Since the data originates from a Polish marketplace, the selection of products reflects items typical to this region. The diversity of products catering to minority groups might be limited due to the popularity-based filtering used in the dataset. Additionally, due to EU regulations, the selection of products may not be representative of other marketplaces, such as those originating from the Americas, Asia, or Africa.

\textbf{E-commerce Domain} Our dataset, composed exclusively of e-commerce product names, may not be easily transferable to the broader NLP domain due to its specialized nature. Product titles often include domain-specific jargon, abbreviations, named entities, numbers, codes, and concise text that differs significantly from the more diverse and unstructured language found in general NLP tasks, such as web pages, articles, or conversations. This domain-specific focus can limit the generalizability of models trained on this data to other NLP applications.
% , highlighting the need for more textual datasets tailored to learning from noisy labels in the scientific community.

\textbf{Machine Translated Content} Product titles have been translated using an in-house Neural Machine Translation (NMT) service, maintained by a team of over 30 machine learning specialists, software engineers, and language quality experts, following recent advancements in NMT. However, machine translation systems, often trained on general language corpora, may struggle with the domain-specific jargon, abbreviations, and structured product descriptions common in e-commerce, leading to inaccurate or misleading translations. Additionally, brand names, model numbers, and industry-specific terms may lack direct equivalents in other languages, resulting in translation errors that can compromise the clarity and reliability of the content. We mitigate these issues through model fine-tuning on in-house data, the use of translation glossaries, input data exceptions, and no-translate entity detection, but the model's accuracy is not perfect. More information on the quantitative impact of machine translations can be found in Appendix~\textbf{\ref{appendix-translation}}.

\textbf{Malicious Content} The product database is maintained daily by expert category managers to detect any malicious behavior on the platform, such as illegitimate, disrespectful, or offensive products, personally identifiable information, derogatory language, etc. To the best of our knowledge, the dataset should be free from malicious content; however, we did not conduct extensive annotation in this regard.

\textbf{Intended Use Case} The intended use case of the dataset is to develop robust text classifiers for benchmarking algorithms that learn from noisy labels, which was our primary focus during its creation. We discourage any unintended usage of the \textit{AlleNoise} dataset.

\textbf{Competing Datasets} To date, several similar benchmark datasets have been published for e-commerce applications of ML algorithms, such as Amazon \cite{hou_bridging_2024}, Rakuten \cite{lin_dataset_2019}, Skroutz \cite{akritidis_self-verifying_2020, akritidis_effective_2018}, and Shopmania \cite{akritidis_self-verifying_2020, akritidis_effective_2018}. Our dataset competes in size (500,000 instances) and content (5,000 categories). A key difference is the known noise level available in the \textit{AlleNoise} dataset.

\section{Impact of title translation on classification performance} \label{appendix-translation}

The dataset was translated from Polish to English, which could impact text representation and the universality of results. To assess this, we compared classifier performance on the original Polish titles versus the translated English ones (\textbf{Tab. \ref{tab:translation}}). The results show that classification accuracy drops by 3.4 p.p. on the clean dataset and by 3.1 p.p. on the noisy dataset after translation, indicating that the translated titles pose a greater challenge for the classifier. However, the goal of the \textit{AlleNoise} dataset is not to maximize classification accuracy but to minimize the performance gap between its noisy and clean variants. The baseline accuracy gap was preserved in translation, amounting to 11.4 p.p. for the Polish version and 11.1 p.p. for the English version. Additionally, the performance of the top-scoring methods on \textit{AlleNoise} (ELR, CCE) showed no significant deviation from the baseline in either version.

% Effect of translation PL->EN
\begin{table}[h]
\centering
\captionsetup{labelfont=bf}
\begin{tabular}{l|cc|cc}
\toprule
& \multicolumn{2}{c|}{Polish} & \multicolumn{2}{c}{English} \\
\midrule
~ & clean & noisy & clean & noisy \\
\midrule
CE & 78.23 ± 0.09 & 66.81 ± 0.13 & 74.85 ± 0.15 & 63.71 ± 0.11 \\
ELR & 78.18 ± 0.07 & 66.94 ± 0.14 & 74.81 ± 0.11 & 63.72 ± 0.19 \\
CCE & 78.17 ± 0.10 & 66.77 ± 0.16 & 74.80 ± 0.09 & 63.73 ± 0.22 \\
\bottomrule
\end{tabular}
\caption{Accuracy of the top-scoring methods on the original Polish and translated English versions of \textit{AlleNoise}. Each score represents the average of 5 cross-validation folds.}
\label{tab:translation}
\end{table}

\section{Relation between classification performance and category size}

\begin{figure}[h]
    \captionsetup{labelfont=bf}
    \centering
    \includegraphics[width=\textwidth]{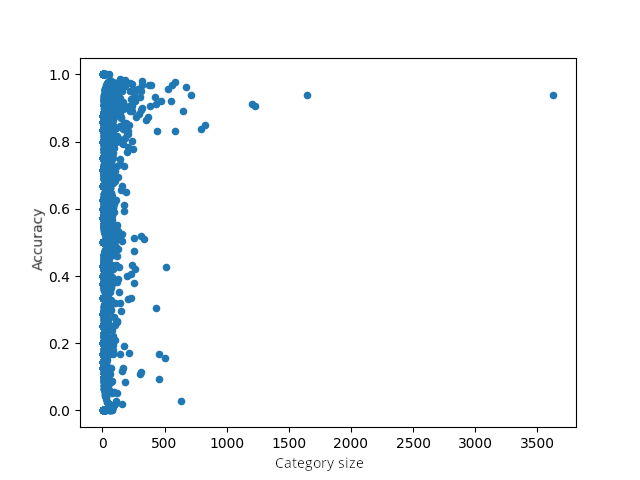}
    \caption{Scatter plot of classification accuracy versus category size for the baseline model using cross-entropy loss. The plot illustrates the lack of correlation between category size and classification accuracy, with a Pearson correlation coefficient of 0.123.}
    \label{fig:imbalance}
\end{figure}

\textit{AlleNoise} is an imbalanced dataset, similar to well-established benchmarks such as Clothing-1M and WebVision. Balancing the classes, as done in CIFAR-10N, poses risks because it may distort category-specific noise levels, making them no longer representative of the original data's noise distribution. While class imbalance could potentially impact performance accuracy, in \textit{AlleNoise}, class size and classification accuracy are uncorrelated (Pearson correlation coefficient: 0.123), as illustrated in \textbf{Fig.~\ref{fig:imbalance}}.

\end{document}